\documentclass{article}
\usepackage{ijcai21}

\usepackage{times}
\usepackage{soul}
\usepackage{url}
\usepackage[hidelinks]{hyperref}
\usepackage[utf8]{inputenc}
\usepackage[small]{caption}
\usepackage{graphicx}
\usepackage{amsmath}
\usepackage{amsthm}
\usepackage{amssymb}
\usepackage{booktabs}
\usepackage{algorithm}
\usepackage{algorithmic}
\usepackage[switch]{lineno}
\usepackage{booktabs}
\usepackage{multirow}

\begin{document}

\pdfinfo{
/TemplateVersion (IJCAI.2021.0)
}

\title{Disentangled Face Attribute Editing via Instance-Aware Latent Space Search}

% Single author syntax
% \author{
%     Paper ID: 1414
% }
% \author{
%     Zhi-Hua Zhou
%     \affiliations
%     Nanjing University
%     \emails
%     pcchair@ijcai-21.org
% }

% Multiple author syntax (remove the single-author syntax above and the \iffalse ... \fi here)
% Check the ijcai21-multiauthor.tex file for detailed instructions
% \iffalse
\author{
Yuxuan Han$^1$
\and
Jiaolong Yang$^2$\footnotemark[2]
\and
Ying Fu$^{1}$\footnotemark[1]
% \thanks{Corresponding Author: fuying@bit.edu.cn}
% Fourth Author$^4$
\affiliations
$^1$Beijing Institute of Technology\\
$^2$Mircosoft Research Asia
% $^3$Third Affiliation\\
% $^4$Fourth Affiliation
\emails
\{hanyuxuan, fuying\}@bit.edu.cn,
jiaoyan@microsoft.com
% fourth@example.com
}
% \fi

% \linenumbers
\maketitle

\renewcommand{\thefootnote}{\fnsymbol{footnote}}
\footnotetext[1]{Corresponding Author: fuying@bit.edu.cn.}
\footnotetext[2]{Work of JY was done in September 2020.}
\renewcommand{\thefootnote}{\arabic{footnote}}

\begin{abstract}
    Recent works have shown that a rich set of semantic directions exist in the latent space of Generative Adversarial Networks (GANs), which enables various facial attribute editing applications.
    However, existing methods may suffer poor attribute variation disentanglement, leading to unwanted change of other attributes when altering the desired one. The semantic directions used by existing methods are at attribute level, which are difficult to model complex attribute correlations, especially in the presence of attribute distribution bias in GAN's training set.
    % In this paper, we propose a novel framework that searches instance-aware semantic directions in the GAN latent space for disentangled attribute editing. 
    In this paper, we propose a novel framework (\textbf{IALS}) that performs \textbf{I}nstance-\textbf{A}ware \textbf{L}atent-Space \textbf{S}earch to find semantic directions for disentangled attribute editing. 
    The instance information is injected by leveraging the supervision from a set of attribute classifiers evaluated on the input images. We further propose a Disentanglement-Transformation ($DT$) metric to quantify the attribute transformation and disentanglement efficacy and find the optimal control factor between attribute-level and instance-specific directions based on it.
    Experimental results on both GAN-generated and real-world images collectively show that our method outperforms state-of-the-art methods proposed recently by a wide margin.
    % \footnote{Code: \url{https://github.com/yxuhan/IALS}}
    Code is available at \url{https://github.com/yxuhan/IALS}.
    %Recent works have shown that a rich set of semantic directions exist in the latent space of Generative Adversarial Networks (GANs) trained on face images, which enables various facial editing applications.
    %Due to the bias of GAN's training set, some attributes are entangled with others in the latent space.
    %However, the semantic direction searched by previous works ignore the instance information and then cannot reflect the entanglement of attributes, which further leads to the failure of disentangled facial editing.
    %In this paper, we propose a novel framework for searching instance-aware semantic direction via the supervision from a Convolution Neural Network (CNN) classifier and the Disentanglement-Transformation ($DT$) metric to evaluate the disentanglement ability of semantic direction based face attribute editing method.
    %Besides, we introduce a control factor to balance the instance-dependent and attribute level information in instance-aware semantic direction and determine it by optimizing the proposed metric.
    %Experimental results show that our method obtain better disentangled face attribute editing results than state-of-the-arts on the synthetic images, and our method can also be extensively implemented on real face editing.
    %%Quality and quantity results show that our method can obtain better disentangled face attribute editing results than previous method. We also extend our framework to real face editing.
\end{abstract}

\section{Introduction}

\begin{figure}[ht]
	\centering
    % \vspace{-10pt}
	\includegraphics[width=0.45\textwidth]{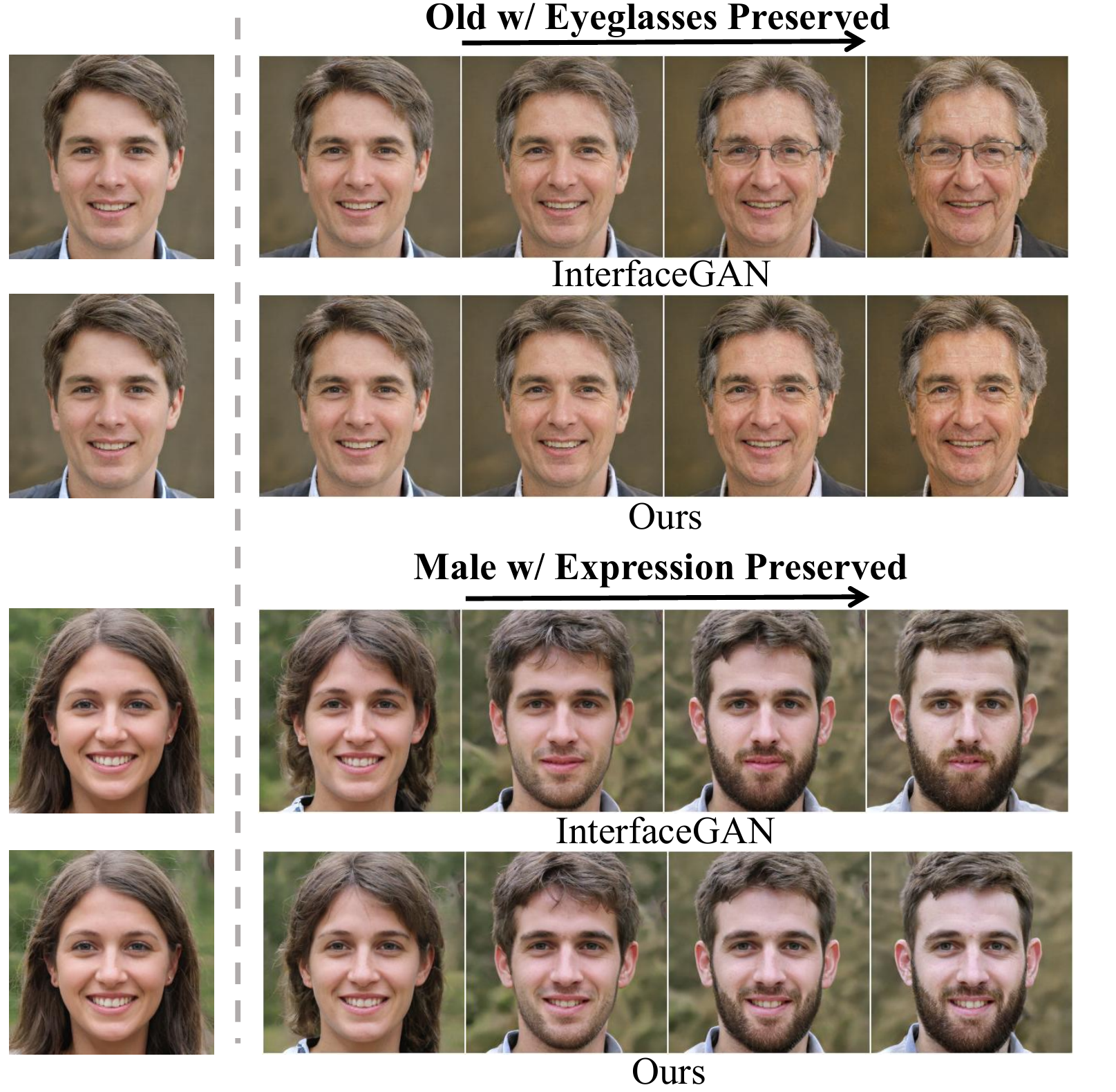} 
	% \includegraphics[width=0.45\textwidth]{img/inconsistency_2.pdf}
	% \vspace{-15pt}
	\caption{Our goal is to change the primal attribute of a given face (\emph{e.g.}, age and gender here) while preserving other condition attributes (\emph{e.g.}, eyeglasses and expression). Despite the latent space directions of the primal and condition attributes in the InterfaceGAN method have been orthogonalized, changing primal attribute still leads to unwanted condition attribute change. Our IALS method can produce satisfactory disentangled editing results.
	%We adopt conditional manipulation~\cite{shen2020interpreting} here, which is to subtract the projection of condition semantic direction from the primal semantic direction. It would work well if the semantic direction can reflect the entanglement between attributes. InterfaceGAN~\cite{shen2020interpreting} failed on this task since the semantic directions of entangled attributes are almost orthogonal to each other, with 0.10 and 0.11 cosine similarity respectively. Our method achieve success since the semantic directions searched by us can reflect the entanglement between attributes, with 0.26 and 0.33 cosine similarity respectively. 
    % \vspace{-10pt}
}
	\label{inconsistency}
\end{figure}
The task of face attribute editing aims to alter a given face image towards a given attribute, such as age, gender, expression, and eyeglasses.
A successful editing should not only output high-quality results with accurate target attribute, but also well preserve all other image content characterized by the complementary attributes.
Face attribute editing has attracted much attention in recent years and numerous algorithms have been  proposed~\cite{shen2017learning,choi2018stargan,zhang2018generative,bahng2020exploring,awiszus2019learning,gu2019mask}. 
%A lot of works have focused on the face attribute editing task in recent years
Notwithstanding the promising results demonstrated by these methods, this task is still quite challenging due to the high output dimension and need for precisely disentangling factors of variation corresponding to different face attributes.
%\ying{please refer to the first paragraph in interFaceGAN and SSCGAN to rewrite this paragraph. }\xuan{ rewrite it refer to SSCGAN}

Face image synthesis has achieved tremendous success recently with  Generative Adversarial Networks (GANs)~\cite{goodfellow2014generative}. State-of-the-art GAN models~\cite{karras2020a,karras2020analyzing} can generate high-fidelity face images from the learned latent space. This motivates several works~\cite{shen2020interpreting,voynov2020unsupervised,shen2020closedform,h2020ganspace} to edit face attribute by reusing the knowledge learned by GAN models. Specifically, for each attribute, they search a corresponding direction in the GAN latent space, such that moving a latent code along this direction can lead to the desired change of this attribute in the generated images. Although the target attributes can be changed effectively by these methods, disentangled editing is still problematic. As illustrated in Figure~\ref{inconsistency}, eyeglasses may appear when age is changed from young to old, despite the latent space direction of age have been made orthogonal to eyeglasses via projection~\cite{shen2020interpreting}. There are at least two possible reasons for this issue: \emph{i}) the distribution bias of GAN's training set (\emph{e.g.}, elder people tend to wear eyeglasses in FFHQ~\cite{karras2020a}), and \emph{ii}) the attribute-level directions cannot handle complex attribute distributions and are not effective for attribute variation disentanglement.

In this paper, we propose a novel framework to search the semantic directions in GAN latent space for disentangled face attribute editing. Instead of naively using fixed, \emph{attribute-level} directions, we opt for dynamically searching \emph{instance-aware} directions, where instance refers to the input image to be edited. 
The intuition behind is that by leveraging the instance information, the complementary attributes of the instance can be explicitly and effectively preserved, leading to disentangled editing results. To render the directions instance-aware, we consider the instance-specific direction obtained by back-propagating the gradient of off-the-shelf attribute CNN classifiers on the input image, and introduce a control factor to balance the attribute-level and instance-specific direction components. 
We propose a Disentanglement-Transformation ($DT$) metric to quantitatively evaluate the editing results and select the control factor that leads to the highest $DT$ metric.
% To study the behavior of different component weighting and seek for an optimal control factor, we propose a Disentanglement-Transformation ($DT$) metric to quantitatively evaluate the editing results. 
% The $DT$ metric considers both primal attribute (\emph{i.e.}, target attribute) editing accuracy (``Transformation") as well as complementary/condition attribute preserving efficacy (``Disentanglement"). 
% We select the control factor that leads to the highest $DT$ metric and use it to edit all input images.

% \subsubsection{GAN Inversion.} 
% The essence of GAN is to learn the mapping function from a given distribution to the real data.
% To manipulate real image leveraging GAN model, we need to invert it back to the latent space, which is also known as GAN inversion.
% Existing works propose to train an encoder~\cite{zhu2019lia} to learn the inverse mapping of GAN, or perform instance level optimization~\cite{abdal2019image2stylegan,abdal2020image2stylegan,gu2020image}.
% Some methods combine these two ideas by leveraging the encoder to output a good initial value for optimization~\cite{zhu2020in}.
% However, this work is orthogonal to the above methods. 
% We use GAN inversion as a tool to enable real face editing in our method.
% Next, we review methods proposed to distill knowledge from the latent space of well-trained GAN.

We test our method on both GAN-generated images and real ones, the latter of which are achieved by GAN inversion \cite{abdal2019image2stylegan} and re-generation. Experiments show that our method can achieve high-quality disentangled face editing results, outperforming state-of-the-art methods on both GAN-generate and real-world images by a wide margin. In summary, our contributions include:
\begin{itemize}
	\item We propose a novel face attribute editing framework that searches for instance-aware semantic direction in GAN latent space, which explicitly promotes attribute variation disentanglement;
	\item We propose a Disentanglement-Transformation ($DT$) metric to quantitatively evaluate the editing efficacy and optimize our algorithm by leveraging this metric;
	\item We achieve high-quality results on both GAN-generated and real images that significantly outperform existing methods.
\end{itemize}

\section{Related Work}
% In this section, we first introduce recent face attribute editing methods. 
% Then, we review the  most related research for our method on semantic direction search in the latent space.

% \subsection{Face Attribute Editing.}
Face attribute editing aims to manipulate the interested face attribute while preserving the rest. 
To achieve this goal, previous methods often leverage the conditional GAN model~\cite{mirza2014conditional,odena2017conditional}.%, which enables various facial editing.
These methods usually design the loss function~\cite{shen2017learning,bahng2020exploring,zhu2017unpaired} or network architecture~\cite{choi2018stargan,zhang2018generative,liu2019stgan,lin2019relgan,he2019attgan} manually to improve the output  quality.
Recently, 3D prior (\emph{e.g.} 3DMM~\cite{blanz1999a}) is also introduced to encode the synthetic image~\cite{deng2020disentangled,tewari2020stylerig}. These methods can generate high-quality results, but the diversity of controllable attributes is limited by the 3D priors (\emph{e.g.}, it cannot well model and edit gender information).
Other recent methods leverage a high-quality synthetic face image dataset for disentangled representation training~\cite{KowalskiECCV2020}.
% Our method edits face attribute by exploring and reusing the knowledge learned by well-trained unconditional GAN model~\cite{karras2018progressive,karras2020a,karras2020analyzing,brock2019large,goodfellow2014generative}. It can edit various face attributes and obtain high-quality results by simply varying the latent code of synthetic image based on the instance-aware semantic direction search. 

% \subsection{Semantic Direction Search in the Latent Space.} 

Another set of works~\cite{shen2020interpreting,voynov2020unsupervised,shen2020closedform,plumerault2020controlling,h2020ganspace,goetschalckx2019ganalyze} propose to edit face attribute by moving the latent code along a specific semantic direction in the latent space of well-trained unconditional GAN model~\cite{karras2018progressive,karras2020a,karras2020analyzing,goodfellow2014generative}.
% Previous works have shown that the latent space of well-trained GAN encodes rich semantics~\cite{yang2019semantic} with vector arithmetic property, and semantic directions can correspond to various face attributes.  
% Face attribute editing can be done by moving the latent code along a specific semantic direction~\cite{shen2020interpreting}.
Shen \emph{et al.}~\cite{shen2020interpreting} learn an SVM boundary to separate the latent space into the opposite semantic label and output the normal vector of the SVM boundary as the semantic direction.
Härkönen \emph{et al.}~\cite{h2020ganspace} sample a collection of  latent codes and perform PCA on them to find principle semantic directions. 
However, these methods search the semantic directions on the attribute level, which cannot handle complex attribute correlations.
Our method dynamically searches instance-aware semantic direction, which is effective for attribute variation disentanglement.
It can edit various face attributes and obtain high-quality results. 

% \subsubsection{GAN Inversion.} 
% The essence of GAN is to learn the mapping function from a given distribution to the real data.
% To manipulate real image leveraging GAN model, we need to invert it back to the latent space, which is also known as GAN inversion.
% Existing works propose to train an encoder~\cite{zhu2019lia} to learn the inverse mapping of GAN, or perform instance level optimization~\cite{abdal2019image2stylegan,abdal2020image2stylegan,gu2020image}.
% Some methods combine these two ideas by leveraging the encoder to output a good initial value for optimization~\cite{zhu2020in}.
% However, this work is orthogonal to the above methods. 
% We use GAN inversion as a tool to enable real face editing in our method.
% Next, we review methods proposed to distill knowledge from the latent space of well-trained GAN.

\begin{figure*}[t]
    \centering
        \includegraphics[width=.82\textwidth]{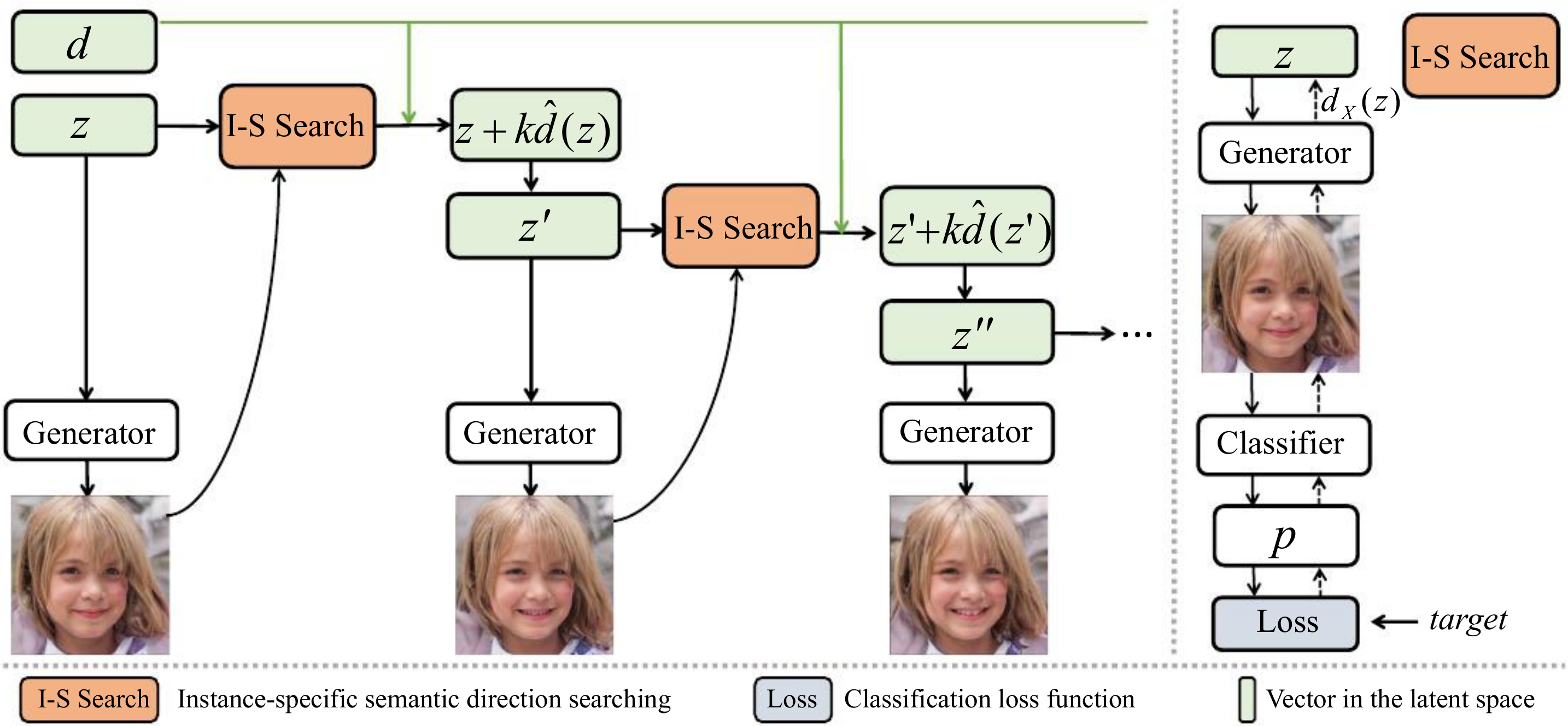}
    \caption{The overview of our face attribute editing framework. The left part shows our instance-aware semantic direction searching method in an 
    incremental update scheme. Here, $d$ and $\hat{d}(z)$ stand for the attribute-level and instance-aware semantic directions, respectively. The right part illustrates the instance-specific semantic direction search process. %\ying{re-plot. Refer to the Figure~ 2 in SSCGAN }\xuan{re-plot it } 
    }
    \label{gradgan_framework}
    % \vspace{-10pt}
\end{figure*}

\section{Method}
This section introduces the novel face attribute editing framework proposed by this paper. We begin by introducing face attribute editing with GANs and briefly revisiting prior methods using attribute-level latent directions, after which we present our instance-aware direction search algorithm.

%In this section, we introduce our framework, which firstly revisits the InterfaceGAN framework. Secondly, we propose a method to search for instance-aware semantic direction. Lastly, we propose an incremental way to edit face attributes with instance-aware semantic direction. \ying{You description should correspond to subsections.}\xuan{rewrite it} %applies it to disentangled face attributes editing.

\subsection{Semantic Direction for Attribute Editing}
Given a pretrained generator $G$ from a state-of-the-art GAN model, \emph{e.g.}, StyleGAN~\cite{karras2020a}, which maps a latent vector $z$ to a face image, attribute editing can be achieved by moving $z$ along a certain direction in the latent space. For real images, one can first embed them into the latent space to obtain the latent vector $z$ and then modify it. The key is to find suitable semantic directions for attribute editing.

Let $\mathcal{A}$ denote  a collection of face attributes, \emph{e.g.}, age, gender, expression, and eyeglasses. 
For each attribute $X\in \mathcal{A}$, existing methods seek for a direction $d_X$ corresponding to $X$. For example, the InterfaceGAN method~\cite{shen2020interpreting} first generates a large corpus of images $G(z)$ by randomly sampling $z$. Then it labels the attributes of these images using a set of CNN binary classifiers $H(\cdot)$. Finally, it trains a SVM to separate each attribute label in the GAN latent space using these samples and outputs the normal vector of the SVM boundary as the semantic direction $d_X$ for each attribute. To achieve disentangled editing, it proposes to edit the primal attribute $A$ while preserving one condition attribute $B$ (or more) via direction orthogonalization:
\begin{equation}
d_{A|B}=d_A-<d_A,d_B>d_B,
\label{cond_mani}
\end{equation}
where $<\cdot,\cdot>$ denotes inner product. Directions with more than one condition attribute can be obtained similarly as discussed in \cite{shen2020interpreting}.

It can be seen that the directions so-obtained are at attribute level. They are fixed for each attribute and are instance-agnostic. We instead propose to incorporate instance information into direction search for better editing performance.

\subsection{Instance-Aware Semantic Direction Search}
Our instance-aware semantic direction consists of two parts: \emph{instance-specific} and \emph{attribute-level} directions. 
Next we first introduce the two components and then describe how to combine them.

\subsubsection{Instance-Specific Semantic Direction}
%The instance-specific semantic direction $d_X(z)$ can only edit the face attribute $X$ of $G(z)$, where $z$ is a \emph{given} latent code in $\mathcal{Z}$. 
%We further assume that the synthetic image of $z+d_X(z)$ possesses the target semantic label of attribute $X$.
The generator $G$ maps a latent code $z$ to an image and an attribute classifier $H$ maps an image $x\in\mathcal{X}$ to an attribute label.
We can bridge the GAN latent space and attribute space via compositing $H$ and $G$, \emph{i.e.} $H(G(\cdot))$. For attribute $X$, we can search for the instance-specific semantic direction for instance $z$, denoted by $d_X(z)$, via minimizing the following loss:
\begin{equation}
    \mathop{\arg\min}\limits_{d_X(z)} L(H(G(z+d_X(z))),y),
    \label{optimization}
\end{equation}
where $y$ is the target attribute label and $L(\cdot,\cdot)$ is the classification loss with binary cross entropy function
\begin{equation}
L(x,y)=-y\log{x}-(1-y)\log(1-x).
\end{equation}

We simply use gradient descent to search for $d_X(z)$ in Eq.~\eqref{optimization}, where $d_X(z)$ is updated by using an incremental direction updating scheme. % in terms of the gradient descent step of Eq.~\eqref{optimization}. 
%We use an incremental direction updating scheme where each time we use one gradient descent step of Eq.~\eqref{optimization} to obtain current $d_X(z)$. 
To streamline the presentation, more details of incremental update is deferred to a later section. We further normalize $d_X(z)$ as the final instance-specific semantic direction for $z$:
\begin{equation}
    \begin{aligned}
        d_X(z) & = \frac{-\nabla_{z} L(H(G(z)),y)}{\|\nabla_{z} L(H(G(z)),y)\|_2} \\
              & = (2y-1)\frac{\nabla_{z} H(G(z))}{\|\nabla_{z} H(G(z))\|_2}.
    \end{aligned}
    \label{onestep}
\end{equation}
Eq.~\eqref{onestep} shows that opposite directions can be obtained with $y=0$ and $1$, respectively. 
%It further indicates that the instance-aware semantic direction for the positive label is just the opposite of its for the negative label.

\subsubsection{Attribute-Level Semantic Direction}
Attribute-level directions aggregate the information across all training instances thus have higher resistance to noise. 
% We leverage the one computed by InterfaceGAN as the default option in our instance-aware direction. 
Similar to previous method, we also leverage attribute-level directions for attribute editing and use the one computed by InterfaceGAN as the default option. 
% though we can compute them in a slightly different fashion.

Here we propose another way to compute the attribute-level direction. The intuition is to bridge the gap between the local instance-level and the global attribute-level information via sampling and averaging. Specifically, we can first randomly sample a set of GAN latent codes $z$ and generate face images by them accordingly. Then, we compute the instance-specific direction for each sample and each attribute according to Eq.~\eqref{onestep}. Finally, for each attribute, we can average the instance-specific directions from all samples as the attribute-level direction $d_X$. We find that these $d_X$ lead to similar attribute editing quality compared to the counterparts computed by InterfaceGAN, but this approach is simpler and easier to implement and eliminates the need for training SVMs.

% Concretely, we first randomly sample a set of GAN latent codes $z$ and generate face images by them accordingly. We then compute the instance-specific direction for each sample and each attribute according to Eq.~\eqref{onestep}. Finally, for each attribute, we average the instance-specific directions from all samples as the attribute-level direction $d_X$. We find that our $d_X$ leads to similar attribute editing quality compared to the counterpart computed by InterfaceGAN, but our process is simpler and easier to implement and we eliminate the need for training SVMs.

\begin{figure}[t]
    \centering
    \includegraphics[width=0.375\textwidth]{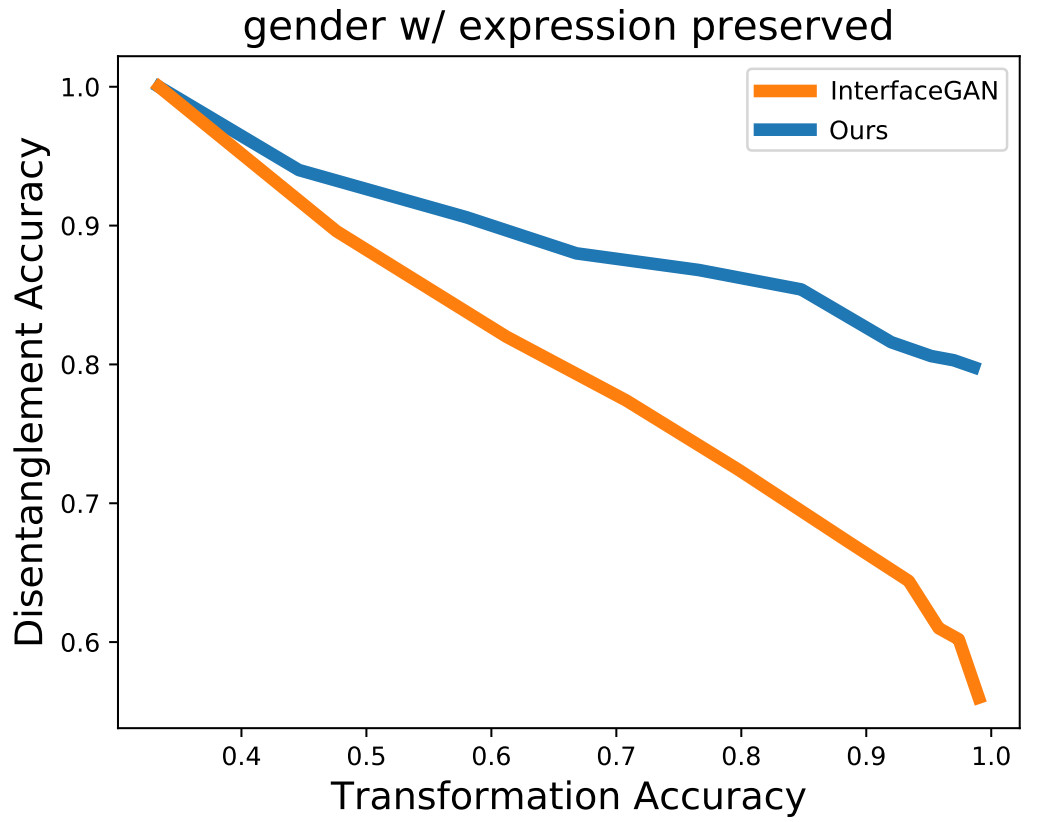}
    % \vspace{-5pt}
    \caption{An example $DT$ curve with gender as the primal attribute and expression as the condition attribute.}
    \label{DTCurve}
    % \vspace{-10pt}
\end{figure}

% \begin{figure}[t]
%     \centering
%     \includegraphics[width=0.4\textwidth, trim=15 15 20 20]{img/DTCurve_jl.pdf}
%     \caption{An example $DT$ curve with gender as the primal attribute and expression as the condition attribute.}
%     \label{DTCurve}
% \end{figure}

\subsubsection{Instance-Aware Semantic Direction}
Our instance-aware semantic direction is constructed by injecting instance-specific information into the direction search process. Specifically, we formulate it  as the combination of the aforementioned attribute-level and instance-specific directions:
\begin{equation}
\hat{d_X}(z)=\lambda d_X + (1-\lambda) d_X(z),
\label{balanced_semantic_direction}
\end{equation}
where $\lambda\in[0,1]$ is the control factor to balance these two components. For conditional attribute editing, we rewrite Eq.~\eqref{cond_mani} with our instance-aware semantic direction as
\begin{equation}
\hat{d_{A|B}}(z)=\hat{d_A}(z)-<\hat{d_A}(z),\hat{d_B}(z)>\hat{d_B}(z),
\label{balanced_cond_mani}
\end{equation}
where $A$ and $B$ are the primal and conditional attributes, respectively.
We set two different values for  $\lambda_1$ and $\lambda_2$ for these two attributes respectively, considering we might require a different quantity of instance information when editing or disentangling face attributes. Note that Eq.~\eqref{balanced_cond_mani} degenerates to Eq.~\eqref{cond_mani} when $\lambda_1 = \lambda_2=1$. In practice, we first solve for the $(\lambda_1,\lambda_2)$ pair which can produce best results on a sample set, after which we fix them for the facial attribute editing task. Next, we discuss how to solve the $(\lambda_1,\lambda_2)$ pair.

%The instance-aware semantic direction is composed of instance-dependent and attribute level semantic direction. 
%To balance the two components, we introduce a control factor $\lambda\in[0,1]$. 
%Then, we formulate the instance-aware semantic direction $\hat{d_X}(z)$ as:
%\begin{equation}
%    \hat{d_X}(z)=\lambda d_X + (1-\lambda) d_X(z)
%    \label{balanced_semantic_direction}
%\end{equation}
%Note that the instance-aware semantic direction can be generalized to the attribute level semantic direction if we set $\lambda=1$.
%Moreover, Eq\eqref{cond_mani} can be reformulated as:
%\begin{equation}
%    \hat{d_{A|B}}(z)=\hat{d_A}(z)-(\hat{d_A}(z),\hat{d_B}(z))\hat{d_B}(z)
%    \label{balanced_cond_mani}
%\end{equation}
%where $A$ is the primal attribute and $B$ is the condition attribute.
%We set two independent $\lambda_1,\lambda_2$ for primal attribute and condition attribute respectively, considering that we might need a different quantity of instance information when editing or disentangling face attributes. 
%Note that Equation~\eqref{cond_mani} is equivalent to Equation~\eqref{balanced_cond_mani} if we set $(\lambda_1,\lambda_2)=(1,1)$.
%In practice, we first solve the $(\lambda_1,\lambda_2)$ pairs which can produce the best results and then complete facial editing with fixed $\lambda_1$ and $\lambda_2$.
%Next, we discuss how to solve the best $(\lambda_1,\lambda_2)$ pair.

As mentioned previously, a good attribute editing should \emph{i}) transform the primal attribute $A$ to target label accurately, and \emph{ii}) preserve the condition attribute $B$ as much as possible. According to these criteria, we propose to use a Disentanglement-Transformation ($DT$) curve to evaluate the editing results for a pair of primal and condition attributes on a set of samples. In a $DT$ curve, the $x$-axis represents the transformation accuracy $p$, \emph{i.e.}, the ratio of samples for which the primal attribute has been transformed into target label correctly in the editing results. The $y$-axis represents the disentanglement accuracy $q$, \emph{i.e.}, the ratio of samples for which the condition attribute on the edited images is consistent with their original ones.

Specifically, we randomly sample a set of latent codes in the GAN latent space and obtain the corresponding images generated by $G$. Then we edit these images by changing the latent code $z$ to $z + k\cdot n\cdot\hat{d_{A|B}}(z)$, where $k\in\mathbb{R}$ is the step size and $n\in\mathbb{N}$ is the number of steps. The transformation and disentanglement accuracy ($p_n$, $q_n$) are evaluated by CNN classifiers for the attributes. By continuously changing $n$ and evaluating ($p_n$, $q_n$), we obtain a $DT$ curve as illustrated in Figure~\ref{DTCurve} (in practice, the $DT$ curve of our method is obtained via varying the incremental updating steps; see next section).
% A better editing algorithm should generate a ``higher" $DT$ curve with higher transformation and disentanglement accuracy. 

With $DT$ curves generated, we can further compute the AUC (Area Under Curve) of $DT$, and we choose the $(\lambda_1,\lambda_2)$ pair that maximizes the average AUC for all possible primal-condition attribute pairs via:
\begin{equation}
\mathop{\arg\max}\limits_{\lambda_1,\lambda_2} \frac{1}{N}\sum_{A,B\in\mathcal{A}} \int ^1_0 q(A,B,\lambda_1,\lambda_2,p) \textup{d}p,
\label{best_lambda}
\end{equation} 
where $N$ is the normalization factor. We numerically estimate the continuous integral in Eq.~\eqref{best_lambda} using quadrature.

\begin{figure}[t]
    \centering 
    \includegraphics[width=0.45\textwidth]{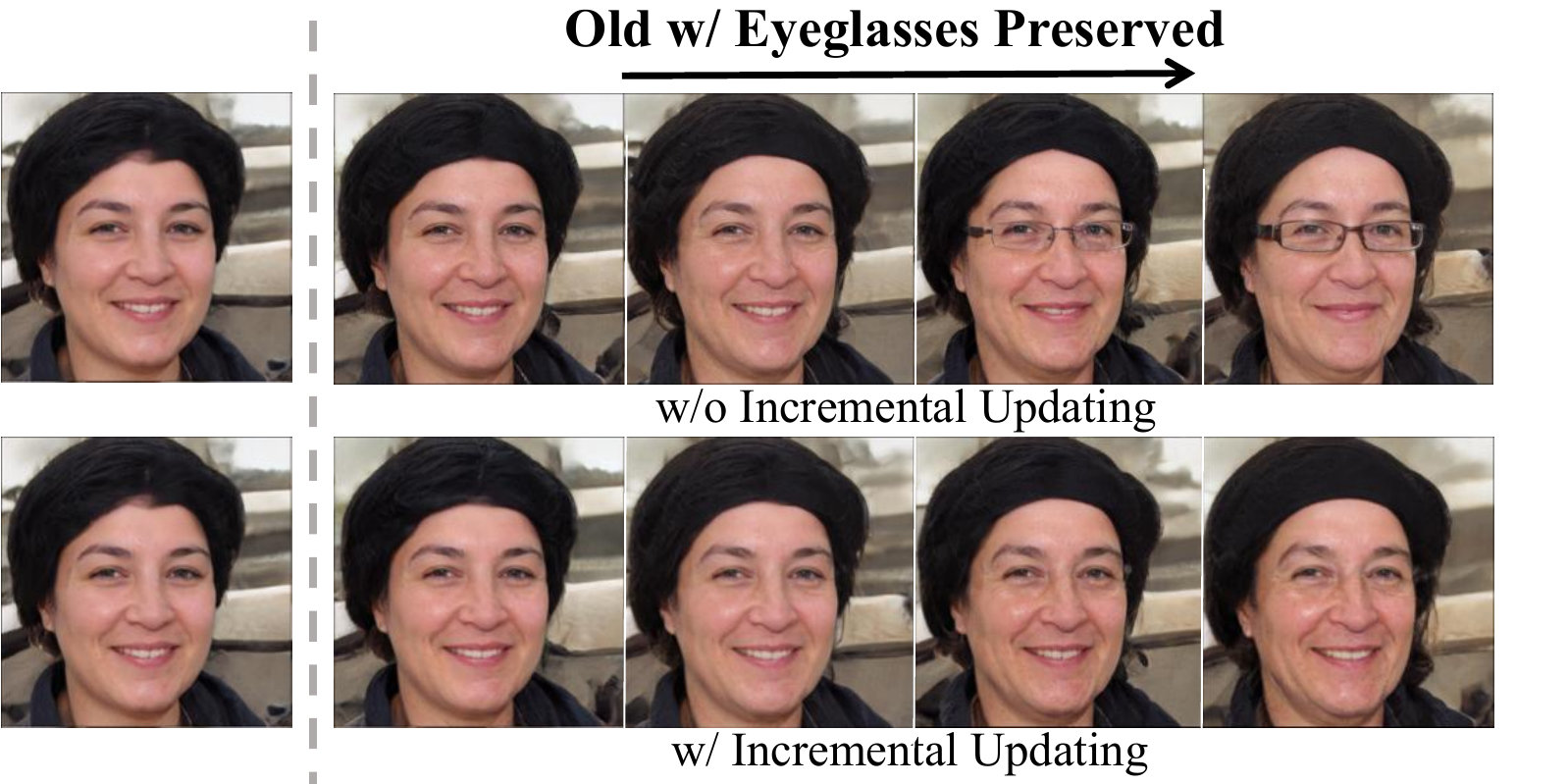}
    % \vspace{-20pt}
    \caption{Ablation study of the incremental updating scheme. }
    \label{inc_update}
    % \vspace{-10pt}
\end{figure}

\begin{figure*}[t]
    \centering
    \includegraphics[width=1\textwidth]{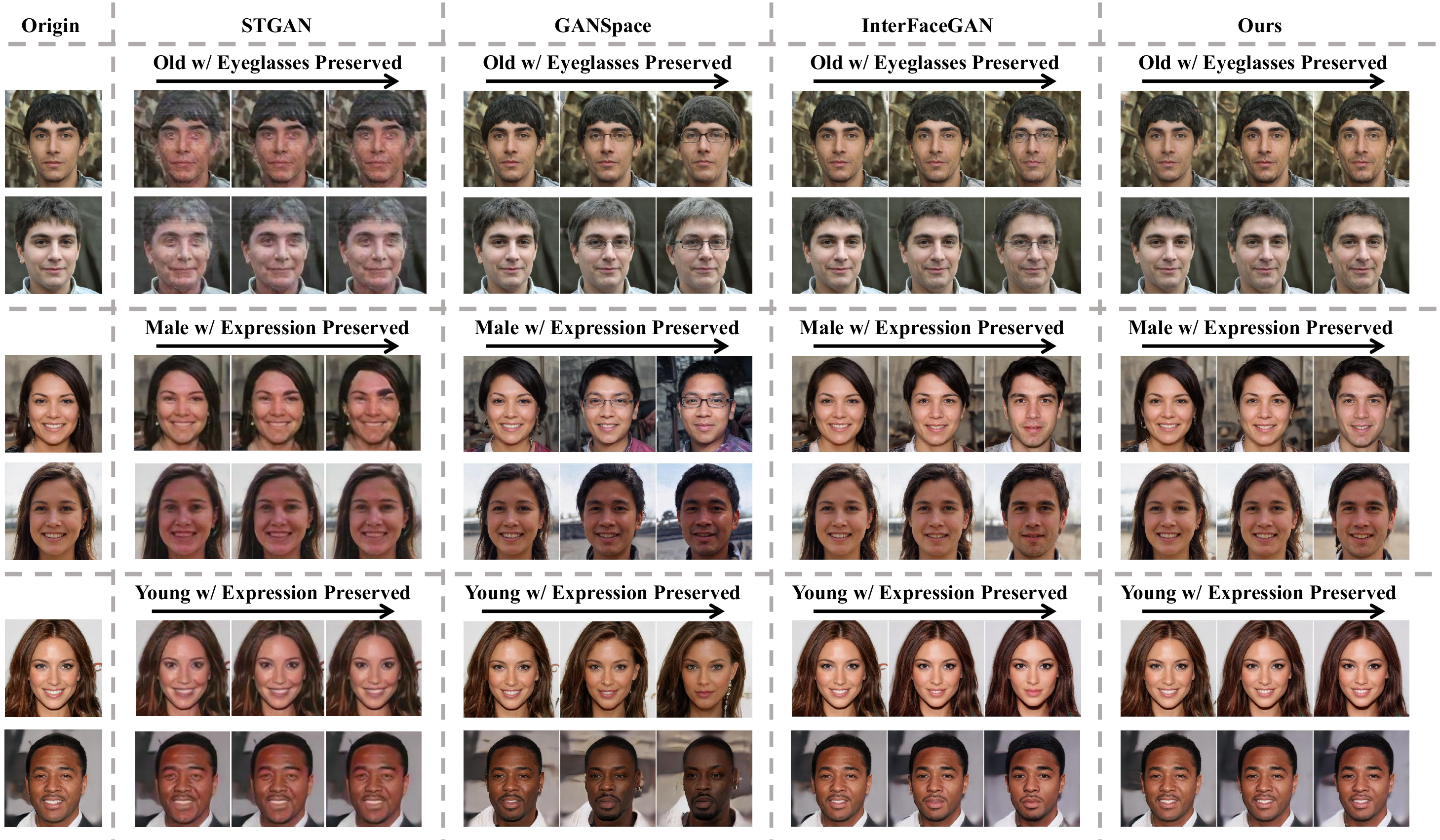}
    \caption{Qualitative comparison of face attribute editing results between our method and other competitors on GAN-generated images. 
    %Our method not only successfully changes the primal attributed as desired but also well preserves the conditional attributes.
    By our method, not only the primal attributes are successfully changed but also the condition attributes are preserved much better than STGAN, GANSpace and InterfaceGAN.}
    \label{compare_to_baseline}
    % \vspace{-5pt}
\end{figure*}

\subsection{Incremental Direction Updating}
Here, we present an incremental instance-aware direction search scheme for our editing method. 
% The intuition behind is that we treat the latent codes formed in the editing process as different instances and use their own instance-aware direction to edit them.
We alternate between searching for the semantic direction and updating the latent code (and the output image) . 
Given the current latent code $z$ and generated image $G(z)$, we search for the new direction $\hat{d}(z)$ and update the latent code as $z' = z + k\cdot\hat{d}(z)$. 
The whole process is illustrated in Figure~\ref{gradgan_framework}. 

\section{Experiment}
% In this section, we evaluate our method and compare it against the state-of-the-arts~\footnote{Our code will be released publicly.}. 
% In this section, we evaluate our method and compare it against the state-of-the-arts. 
In this section, we evaluate our IALS method and compare it against the state-of-the-arts. 
Due to space limitation, more experimental results and discussions are included in the \emph{suppl. material}.
%In this section, we first provide the settings in our experiment. 
%Then, we perform ablation studies to evaluate how can different choices affect the results. 
%Finally, we compare our method to state-of-the-arts by conducting qualitative and quantitative experiments.   

\subsection{Settings and Implementation Details}
We consider five face attributes in our experiment, \emph{i.e.}, expression, age, gender, eyeglasses, and pose, as in \cite{shen2020interpreting}. We will focus on the former four attributes in our experiments as we find pose is well disentangled from other attributes in GAN latent space (similar observations are found in \cite{shen2020interpreting,deng2020disentangled}). We test our framework on the $\mathcal{W}$ space of StyleGAN generator trained on the FFHQ~\cite{karras2018progressive} and CelebA-HQ~\cite{karras2020a} datasets. 
The attribute classifiers $H(\cdot)$ are ResNet-18~\cite{he2016deep} networks trained on the CelebA dataset~\cite{liu2015deep}. 
The $DT$ metrics are computed by another set of attribute classifiers with ResNet-50 structure.
We empirically set the step size of incremental updating in our method to $0.1$ in the following experiments.

\subsection{Ablation Study}
In this part, we investigate the effect of direction control factor and incremental updating scheme in our IALS framework.

\subsubsection{Control Factor Pairs}\label{sec:test}
To study the behavior of different combination weights for attribute-level and instance-specific directions, and solve for control factors $\lambda_1$ and $\lambda_2$, we evaluate the $DT$ metric with $\lambda_1$ and $\lambda_2$ evenly sampled in $[0,1]$ with a step size of 0.25 (hence 25 ($\lambda_1$, $\lambda_2$) pairs in total). 
For each ($\lambda_1$, $\lambda_2$) pair, we set $n_{max}=20$ (\emph{i.e.} sample 20 points on the $DT$ curve) and $k=0.1$ and adopt the trapezoidal quadrature formula to approximate the integral in AUC computation defined in Eq.~\eqref{best_lambda}.

%To solve for optimal control factor pair, we do not solve Equation~\eqref{best_lambda} directly, instead, we relax it by the following settings: 1) we search $\lambda_1$ and $\lambda_2$ discretely with a step size of 0.25, \emph{i.e.} we consider 25 $(\lambda_1,\lambda_2)$ pairs uniformly in $[0,1]\times[0,1]$ 2) we estimate the $DT$ curve by sampling 20 points on it and adopt the trapezoidal quadrature formula to approximate the integral of $DT$ in $[0,1]$. 
%In addition, we use the semantic direction searched by InterfaceGAN as the attribute level component in instance-aware semantic directions.

Table~\ref{auc} shows the average AUC of all $DT$ curves with different primal-condition attribute combinations. We have the following three observations: \emph{i}) using attribute-level direction alone (\emph{i.e.,} $\lambda_1 =\lambda_2 = 1$) is clearly non-optimal; \emph{ii}) the attribute-level information is more important to edit primal attribute ($\lambda_1\geq 0.75$); and \emph{iii}) adding instance information to condition attribute direction search significantly improves the editing results ($\lambda_2\leq 0.75$).

\begin{table}[]
    \centering
    \begin{tabular}{cccccc}
        \hline
        \multicolumn{1}{l}{$\lambda_1$\textbackslash{}$\lambda_2$} & 0.0             & 0.25            & 0.5    & 0.75   & 1.0    \\ \hline
        0.0                                                          & 0.8393          & 0.8454          & 0.8445 & 0.8444 & 0.8302 \\ 
        0.25                                                         & 0.8717          & 0.8675          & 0.8684 & 0.8706 & 0.8463 \\ 
        0.5                                                          & 0.8882          & 0.8889          & 0.8886 & 0.8922 & 0.8660 \\ 
        0.75                                                         & \textbf{\emph{0.9025}} & \emph{0.8992} & \emph{0.9010} & \emph{0.9003} & 0.8742 \\ 
        1.0                                                          & \emph{0.8999} & \emph{0.9003} & \emph{0.9006} & \emph{0.9013} & 0.8662 \\ \hline
    \end{tabular}
    
    \caption{The average AUC (Area Under Curve) of all $DT$ curves with different $(\lambda_1,\lambda_2)$ choices. }
    \label{auc}
    % \vspace{-5pt}
\end{table}

\paragraph{Discussion}
Table \ref{auc} shows a performance plateau for a range of $\lambda$'s ($\lambda_1\geq 0.75$, $\lambda_2\leq 0.75$), indicating that our method is insensitive to parameter selection within a reasonable range. It also shows that adding moderate instance-level information for the condition attribute would significantly better than using attribute-level information alone ($\lambda_2 = 1$), demonstrating the efficacy of our framework.
% In the following experiments, we simply use $(\lambda_1,\lambda_2)=(0.75,0)$ for our editing method. 
In the following we simply use $(\lambda_1,\lambda_2)=(0.75,0)$ for our editing method. 

%We find that the combination weights corresponding to the italic items in Table~\ref{auc} lead to similar results under the $DT$ metric.
%However, we want to emphasize that all of them are significantly better than the result when using attribute-level direction alone.   
%It demonstrates the efficacy of the combination of instance and attribute-level information. 

%As illustrated in Table~\ref{auc}, we obtain the optimal results when  $(\lambda_1,\lambda_2)=(0.75,0)$.
%It indicates that the attribute level information is more important to edit the primal attribute while the instance-dependent information is more important to disentangle condition attribute.
%Furthermore, it is necessary to set two independent control factor for primal and condition attributes respectively, for the optimal performance would be decreased if $\lambda_1=\lambda_2$.

\subsubsection{Incremental Updating}
We further study the efficacy of our incremental updating scheme for instance-aware direction search. 
Figure~\ref{inc_update} shows a typical result when changing age from young to old while preserving the eyeglasses attribute of the original face image. It can be seen that the results are clearly inferior if we do not use incremental updating (\emph{i.e.}, we keep using the instance-aware direction obtained in the first iteration).
%We can see that if we do not adopt the incremental updating, we still fail to disentangle the condition attribute from the primal attribute.

\begin{figure*}[t]
% \vspace{10pt}
    \centering
    \includegraphics[width=1\textwidth]{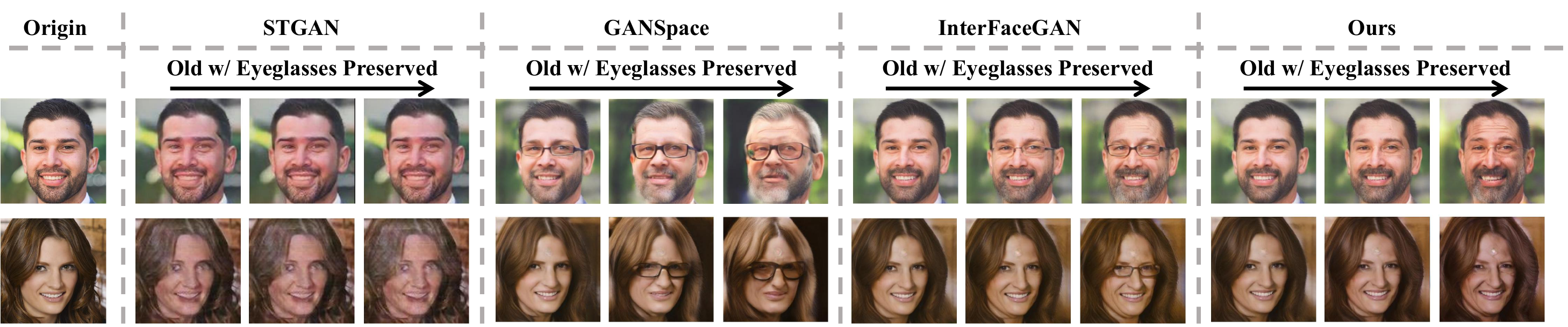}
    % \caption{Qualitative comparison of face attribute editing results between our method and other competitors on real images taken from FFHQ / CC BY-NC-SA $4.0$. Our method yields higher fidelity results than STGAN and better attribute variation disentanglement than GANSpace and InterfaceGAN. }
    % \label{real}
    \caption{Qualitative comparison of face attribute editing results between our method and other competitors on real images from FFHQ. Our method yields higher fidelity results than STGAN and better attribute variation disentanglement than GANSpace and InterfaceGAN. Photos by Flickr users Elio Yañez (Creative Commons BY 2.0 License) and sexinhose (Public Domain Mark 1.0 License).}
    \label{real}
    % \vspace{0pt}
\end{figure*}

\subsection{Comparison with the State-of-the-arts}
We compare our IALS method with several state-of-the-art face attribute editing method proposed recently, including InterfaceGAN~\cite{shen2020interpreting}, GANSpace~\cite{h2020ganspace}, and STGAN~\cite{liu2019stgan}.% conditional GAN based methods. 

%We firstly briefly introduce these methods, then conduct experiments on synthetic images, and lastly on real face images.   

%, including semantic direction~\cite{shen2020interpreting, h2020ganspace} and conditional GAN based methods~\cite{liu2019stgan}. 

Both InterfaceGAN and GANSpace are methods based on GAN latent space search. InterfaceGAN is a supervised semantic direction search method as mentioned in the previous section, while GANSpace is an unsupervised one which performs PCA on the sampled latent codes to find principle semantic directions in the latent space. We assign the directions found by GANSpace to interpretable meanings following~\cite{shen2020closedform}. The STGAN method is based on image generation using conditional GAN. It uses an encoder-decoder architecture with a well-designed selective transfer unit for attribute editing. 
For these methods, we use the code or pretrained models released by the authors for evaluation and comparison.

%\subsubsection{Compared Methods.}
%For the semantic direction based methods, InterfaceGAN~\cite{shen2020interpreting} and GANSpace~\cite{h2020ganspace} are used as the competing approaches. 
%Here, InterfaceGAN is a supervised method as mentioned in the previous section while GANSpace is an unsupervised method. 
%Specifically, GANSpace performs PCA on the sampled latent codes to find principle semantic directions in the latent space. 
%In addition, we assign the semantic directions searched by GANSpace to interpretable meanings following~\cite{shen2020closedform}. 
%For the conditional GAN based method, we adopt the state-of-the-art STGAN~\cite{liu2019stgan} model, which uses the encoder-decoder architecture with a well-designed selective transfer unit. 

%\ying{The pretrained model is trained on the same dataset with ours?} 

%\begin{table}[t]
%    \centering
%    \begin{tabular}{ccc}
%        \hline
%        GANSpace & InterfaceGAN &  ~~ Ours~~   \\ \hline
%          0.7789  & 0.8662 &\textbf{0.9025} \\ \hline
%    \end{tabular}
%    \caption{The average AUC (Area Under Curve) of $DT$ curves from methods that based on GAN latent space search.}
%    \label{quantity_compare}
%\end{table}

\subsubsection{Qualitative Results}\label{sec:qualitative}
we compare our method with the competitors on synthetic images and real images respectively.
We adopt the conditional manipulation setup~\cite{shen2020interpreting} with one primal attribute to be changed and one condition attribute to be preserved for the semantic direction based methods, and directly send the source image to the conditional-GAN based method. 

Some typical qualitative results on GAN-generated images are provided in Figure~\ref{compare_to_baseline}. 
In can be seen that as the degree of attribute change increases, STGAN generated blurry images while the semantic direction based methods outputted high-fidelity results. 
Furthermore, our method can well preserve the condition attribute of the original image when editing the primal attribute while the other two semantic direction based competitors often fail.
InterfaceGAN and GANSpace only use attribute-level direction while ignoring the instance information when facial editing.
By contrast, our method combines instance-specific and attribute-level information which results in much better disentangled facial editing results. 
The results also demonstrate that our method outperforms state-of-the-art methods on GAN-generated images.

Figure~\ref{real} shows two typical examples to illustrate the attribute editing efficacy of different methods on real images\footnote{For InterfaceGAN, GANSpace and our method, we firstly embed the given images into the $\mathcal{W}+$ latent space of StyleGAN using \cite{abdal2019image2stylegan} and then edit them.} where the goal is to edit age while preserving the eyeglasses attribute. 
We find that STGAN outputted blurry results again.  
On the other hand, the condition attributes (\emph{i.e.} eyeglasses) are changed by GANSpace and InterfaceGAN. 
In contrast, our method obtains superior results in terms of both image quality and attribute variation disentanglement. 
It produces high fidelity results with satisfactory attribute modification and preservation.

% \begin{table}[]
%     \centering
%     \begin{tabular}{@{}cccc@{}}
%     \toprule
%     \multirow{2}{*}{Method} & \multicolumn{3}{c}{Human Satisfactory Rate}            \\ \cmidrule(l){2-4} 
%                             & Sup. Pairs  & Other Pairs & All Pairs       \\ \cmidrule{1-4}
%     STGAN                   & 3.63\%           & 7.28\%           & 6.48\%           \\
%     GANSpace                & 40.68\%          & 33.97\%          & 35.44\%          \\
%     InterfaceGAN            & 57.87\%          & 63.24\%          & 62.06\%          \\
%     Ours                    & \textbf{77.00\%} & \textbf{67.60\%} & \textbf{69.66\%} \\ \bottomrule
%     \end{tabular}
%     \caption{User study results over competitors on different set of primal-condition attribute pairs.}
%     \label{quant}
% \end{table}

\begin{figure}[]
    % \vspace{-10pt}
    \centering
    \includegraphics[width=0.45\textwidth]{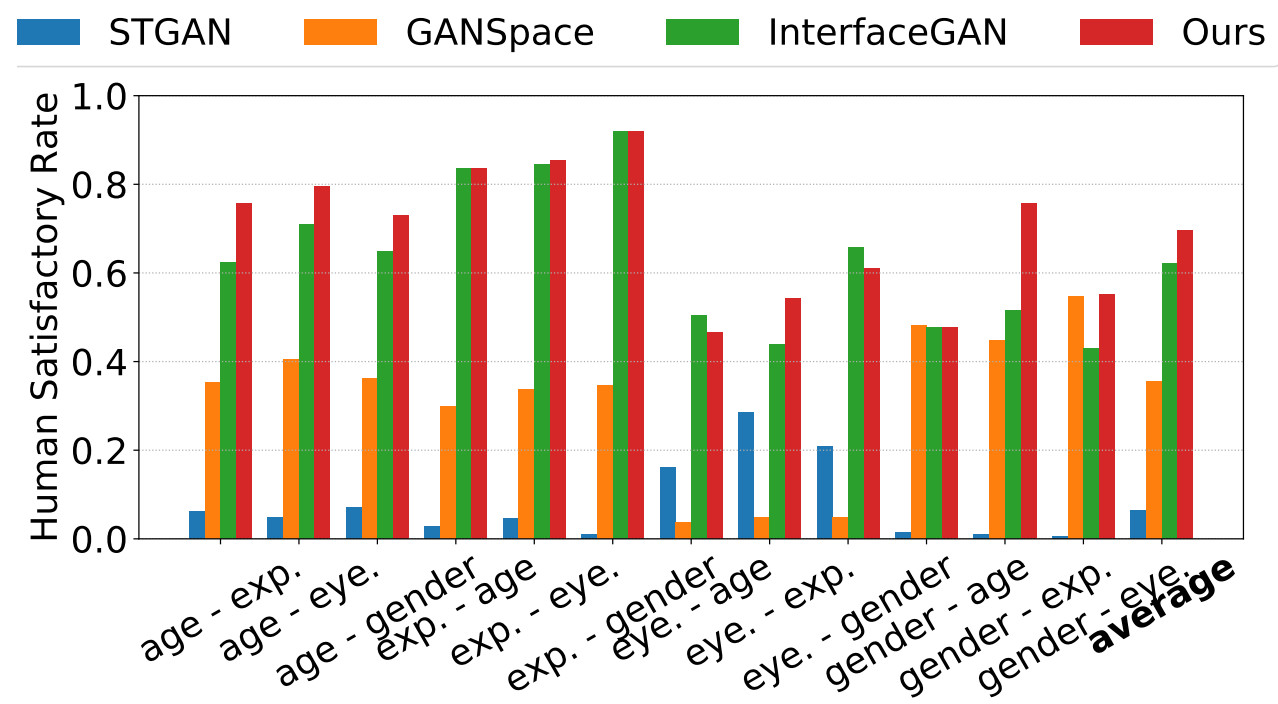}
    % \vspace{-8pt}
    \caption{Human satisfactory rate comparison between our method and other competitors on different primal-condition attribute pairs (exp. and eye. denote expression and eyeglasses respectively)}
    \label{quant}
    % \vspace{-12pt}
\end{figure}

% \begin{figure}[]
%     \vspace{-10pt}
%     \centering
%     \includegraphics[width=0.49\textwidth, trim=45 5 50 50,clip]{img/rescol111.pdf}
%     \caption{Human satisfactory rate comparison between our method and other competitors on different primal-condition attribute pairs (exp. and eye. denote expression and eyeglasses respectively)}
%     \label{quant}
%     \vspace{-12pt}
% \end{figure}

\subsubsection{Quantitative Results}
To further evaluate the performance of our IALS method, we conducted a user study. 
we recruited 100 people and presented them with 720 groups of data, with each group consisting of 5 images - the original face image
% \footnote{All of the original face images randomly are generated by StyleGAN trained on FFHQ or CelebAHQ.} 
and the facial editing results of our method and the other 3 competitors.
% \footnote{We adopt the same implementation setup described in the previous qualitative experiment. }.
Each person was randomly assigned 18 groups of data and asked to choose all of the results they are satisfied with according to three criteria: the result looks natural, the primal attribute is well changed, and condition attribute is well preserved. 
Note that we did not ask them to select the best result, as optimal facial editing results are often not unique.

The results in Fig.~\ref{quant} show that our method obtained the highest average satisfactory rate (69.66\% for ours vs. 62.06\% for InterfaceGAN, 35.44\% for GANSpace and 6.48\% for STGAN). It significantly outperforms the other methods on some challenging attribute pairs such as gender as the primal attribute and expression as the condition attribute (76.44\% for ours vs. 51.92\% for InterfaceGAN).
%It confirms the qualitative results shown in the previous section.

% We denote the \emph{gender-expression} and \emph{age-expression} pairs together as Sup. Pairs and the rest primal-condition pairs as Other Pairs. The results in Table~\ref{quant} indicate that \emph{i)} our method achieved significantly better results on Sup. Pairs; \emph{ii)} our method obtained the highest average satisfactory rate over the competitors on all of the primal-condition attribute pairs.
% It confirms the qualitative results shown in the previous section.
% Table~\ref{quant} shows that our method obtained the highest average satisfactory rate over the competitors on all of the primal-condition attribute pairs.
% It confirms the qualitative results shown in the previous section.
% Besides, we notice that our method achieve significantly better results on the \emph{gender-expression} and \emph{age-expression} pairs which are denoted together as Adv. Pairs. 

\section{Conclusion}
We have proposed a novel IALS framework to achieve disentangled face attribute editing with GAN latent space search. The key ingredient is the integration of attribute-level and instance-specific directions, which leads to accurate target attribute control and, more importantly, highly disentangled attribute editing. To better integrate the attribute-level and instance-specific directions, we introduce a Disentanglement-Transformation ($DT$) metric to find suitable control factor. The experiments collectively show that our approach obtains significantly better results than previous methods. 

\section*{Acknowledgements}
This research was supported by the National Natural Science Foundation of China under Grants No. 61827901 and No. 62088101.
%instance-aware semantic direction search in the GAN latent space.  and the $DT$ metric to evaluate the disentanglement face attribute editing effect. 
%We introduce instance-dependent information to the semantic direction, which is ignored in existing works. 
%We search for the best control factor to balance the instance-dependent and attribute level information in the semantic direction by optimizing the proposed metric. 
%Experimental results show that our method obtains better disentanglement facial editing results than  state-of-the-arts, and we also apply our method to real face manipulation.
% \section{Acknowledgments}
% This research was funded by the National Natural Science Foundation of China under Grants No. 61827901 and No. 62088101.

\part*{Supplementary Material}

\section{More Implementation Details}
We consider five attributes in our implementation, \emph{i.e.}, expression, age, gender, eyeglasses and pose, as in \cite{shen2020interpreting}. 
Note that our method can be extended to edit any other attributes in the presence of corresponding CNN classifier.
%if the corresponding CNN binary classifier is available.
To search instance-specific directions, we train a set of binary CNN classifiers $H(\cdot)$ on the CelebA dataset~\cite{liu2015deep} with $256\times256$ image resolution. 
Unlike other attributes, pose is not annotated in CelebA. We therefore construct the training set for it by ourselves.
%The pose attribute is not annotated in CelebA like other attributes we consider, \emph{e.g.}, gender. Thus, we need to construct the training set for it by ourselves. 
Specifically, we first use the landmarks provided by CelebA annotations to estimate the Euler angles of the face images, and then label images with yaw angle greater than $30^{\circ}$ as a positive sample and less than $-30^{\circ}$ as a negative sample, respectively. %\ying{since it is not a labeled attribute in CelebA like gender?}\xuan{rewrite}
In the training phase, we first center-crop the original $178\times218$ images and then resize them to $256\times256$ before feeding them to $H(\cdot)$ (we adopt the ResNet-18~\cite{he2016deep} architecture), similar to other attributes. Table~\ref{CNNres} shows the precision and recall of our trained classifiers for all attributes.
%As shown in Table~\ref{CNNres}, our trained classifiers achieves high classification precision and recall. 

For instance-specific direction searching, we resize the images generated by $G(\cdot)$ (\emph{i.e.}, the StyleGAN~\cite{karras2020a} generator) from $1024\times1024$ to $256\times256$
%by average pooling 
before feeding them to $H(\cdot)$. 
% The pseudocode can be found in Algorithm~\ref{Instance-Specific Direction Search}.
In the instance-aware direction constructing process, we first normalize the instance-specific direction $d_X(z)$ and the attribute-level direction $d_X$ to unit length, and then combine them % to unit length before combining them
by the control factor $\lambda$.
In addition, we normalize $\hat{d}_X(z)$ as the final instance-aware semantic direction for $z$ on attribute $X$.
% In the incremental updating scheme, we empirically set the step size to $k=0.1$.
% The pseudocode of our method is provided  in Figure~\ref{Pseudocode}.
% The pseudocode can be found in Algorithm~\ref{Instance-Aware Direction Search}. Besides, we provide the pseudocode of incremental updating in Algorithm~\ref{incremental update}.

The pseudocode of our method is provided in Algorithm~\ref{our method}.

\begin{table*}[t]
    \centering
    \begin{tabular}{cccccc}
        \hline
                 & Smiling & Gender & Age  & Eyeglasses & Pose \\ \hline
        Precision & 0.9407  & 0.9613 & 0.9609 & 0.9951  & 0.9884  \\
        Recall   & 0.9094  & 0.9855 & 0.8593 & 0.9264  & 0.9906  \\ \hline
    \end{tabular}
    \caption{Performance of the CNN attribute classifiers (ResNet-18) trained on CelebA. These classifiers are used to edit attributes in our method.}
    \label{CNNres}
    \vspace{5pt}
\end{table*}

% \begin{algorithm}[h]
%     \SetKwData{Left}{left}\SetKwData{This}{this}\SetKwData{Up}{up}
%     \SetKwFunction{Union}{Union}\SetKwFunction{FindCompress}{FindCompress}
%     \SetKwInOut{Input}{input}\SetKwInOut{Output}{output}
%     \Input{latent vector $z$, StyleGAN generator $G$, CNN classifier $H$, attribute-level direction $d_X$, target label $y$, control factor $\lambda$, step size $k$, number of step $N$}
%     \Output{a list $\mathcal{I}$ containing $N$ images with continuous attribute variation towards the target label.}
%     $i\leftarrow 0, ~ \mathcal{I}\leftarrow \emptyset$ \\
%     \While{$i<N$}{
%         $d_X(z)=\frac{\partial L(H(G(z)),y)}{\partial z}$ ~(Eq. 4 in the main paper)\\
%         $d_X(z)\leftarrow Normalize(d_X(z))$\\
%         $d_X\leftarrow Normalize(d_X)$\\
%         $\hat{d}_X(z)=\lambda d_X+(1-\lambda)d_X(z)$ ~~(Eq. 5)\\
%         $\hat{d}_X(z)=Normalize(\hat{d}_X(z))$\\
%         % $\hat{d}_X(z)\leftarrow InstanceAwareDirectionSearch(G,H,L,d_X,z,y,\lambda)$\\
%         $z\leftarrow z+k\cdot\hat{d}_X(z)$\\
%         $I\leftarrow AppendList(I,G(z))$\\
%         $i\leftarrow i+1$
%     }
%     \caption{Instance-Aware Semantic Direction Search for Face Attribute Editing}
%     \label{our method}
% \end{algorithm}

\begin{algorithm}[]
    \caption{Instance-Aware Semantic Direction Search for Face Attribute Editing}
    \label{our method}
    \textbf{Input}: latent vector $z$, StyleGAN generator $G$, CNN classifier $H$, attribute-level direction $d_X$, target label $y$, control factor $\lambda$, step size $k$, number of step $N$. \\
    \textbf{Output}: a list $\mathcal{I}$ containing $N$ images with continuous attribute variation towards the target label.
    \begin{algorithmic}[1] %[1] enables line numbers
        \STATE Let $i\leftarrow 0, ~ \mathcal{\mathcal{I}}\leftarrow \emptyset$.
        \WHILE{$i<N$}
        \STATE $d_X(z)=\frac{\partial L(H(G(z)),y)}{\partial z}$ ~(Eq. 4 in the main paper)\\
        \STATE $d_X(z)\leftarrow Normalize(d_X(z))$\\
        \STATE $d_X\leftarrow Normalize(d_X)$\\
        \STATE $\hat{d}_X(z)=\lambda d_X+(1-\lambda)d_X(z)$ ~~(Eq. 5)\\
        \STATE $\hat{d}_X(z)=Normalize(\hat{d}_X(z))$\\
        \STATE $z\leftarrow z+k\cdot\hat{d}_X(z)$\\
        \STATE $\mathcal{I}\leftarrow AppendList(\mathcal{I},G(z))$\\
        \STATE $i\leftarrow i+1$
        
        \ENDWHILE
        \STATE \textbf{return} $\mathcal{I}$
        \end{algorithmic}
\end{algorithm}

\section{More Results and Comparisons}

\subsection{More Results of Our Method} 
\begin{figure*}[t]
    \centering
    \includegraphics[width=0.95\textwidth]{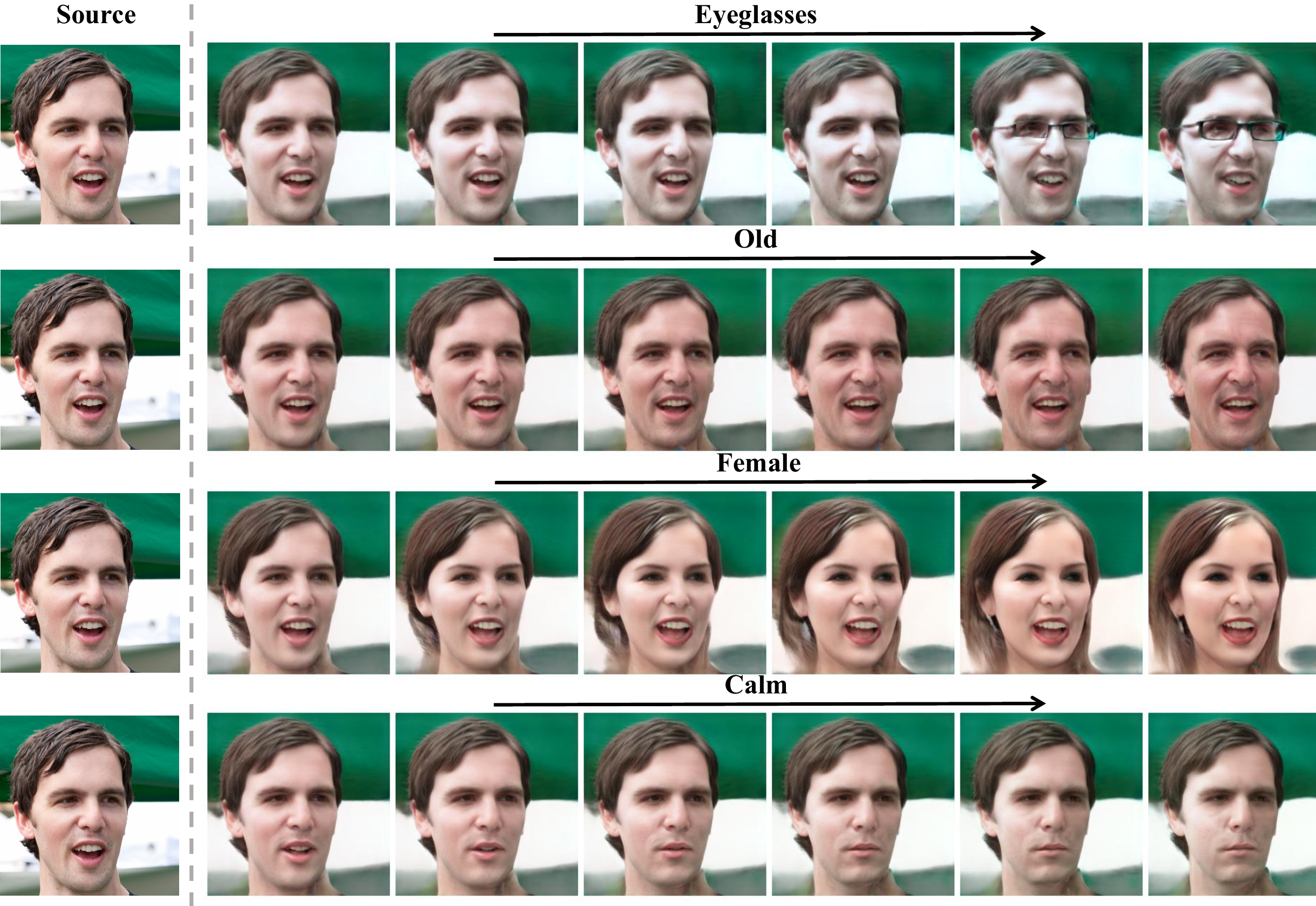}
    \includegraphics[width=0.95\textwidth]{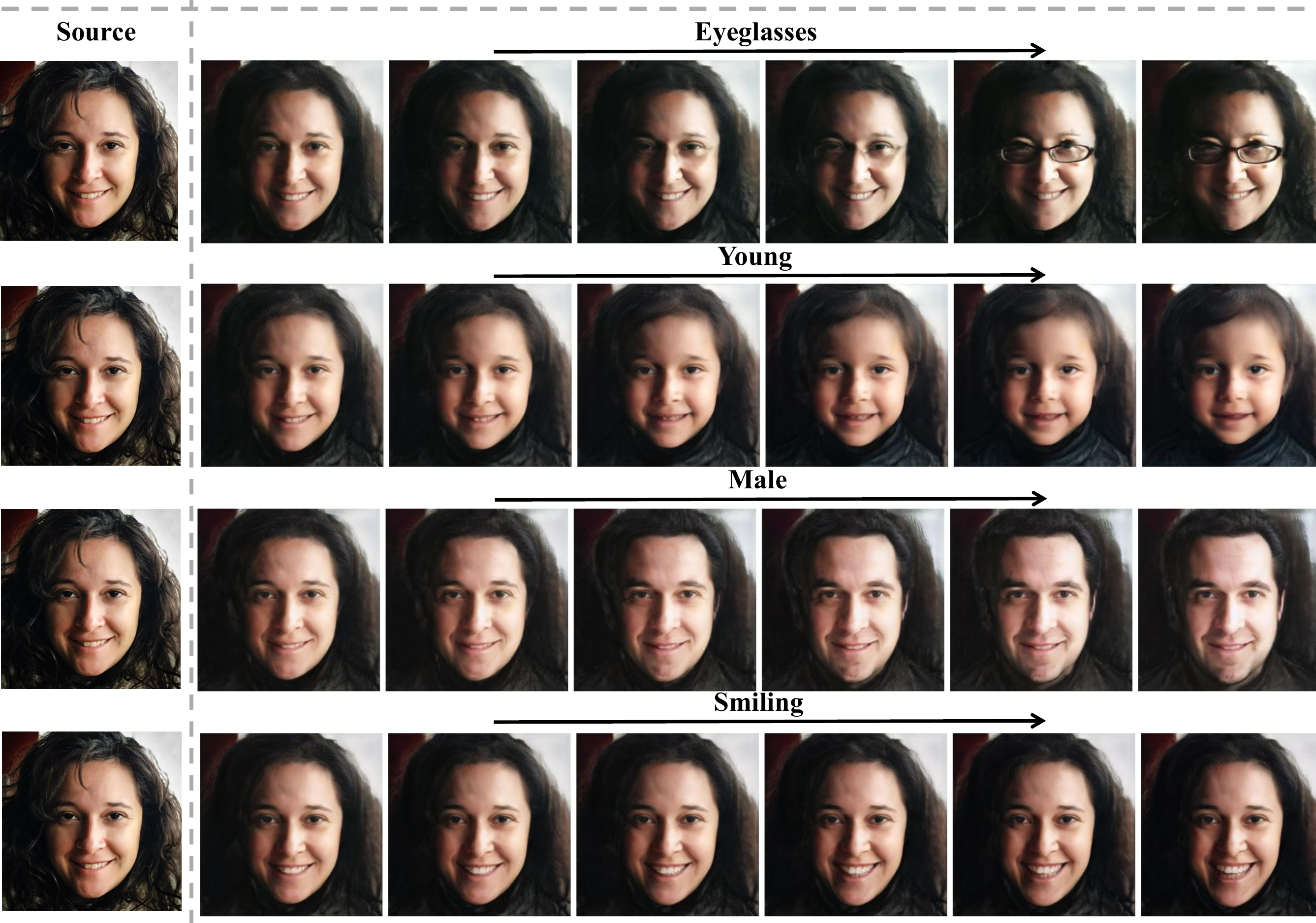}
    \caption{More qualitative results of our method on real image editing. The images are taken from FFHQ.}
    \label{real_1}
\end{figure*}

\begin{figure*}[t]
    \centering
    \includegraphics[width=0.95\textwidth]{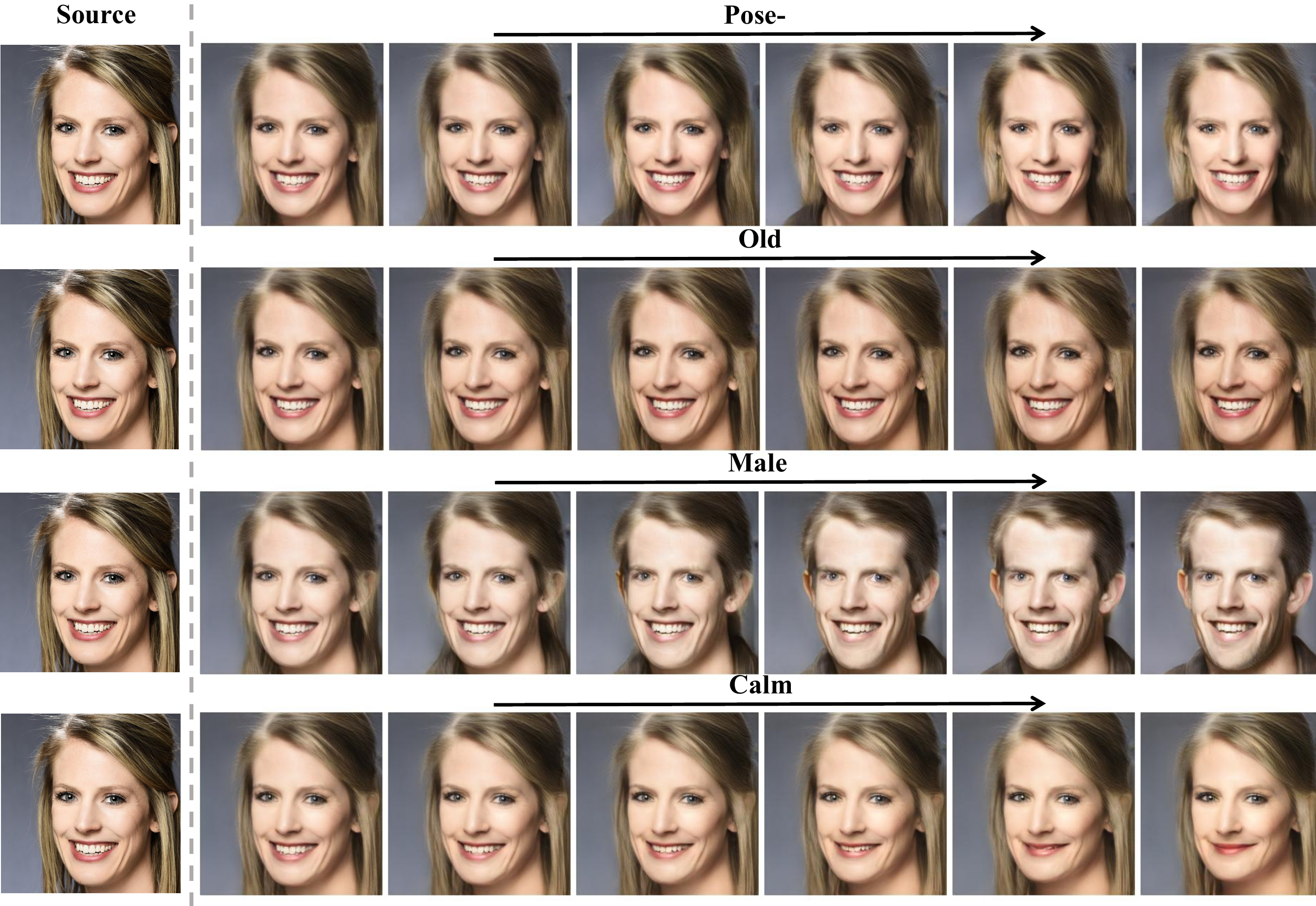}
    \includegraphics[width=0.95\textwidth]{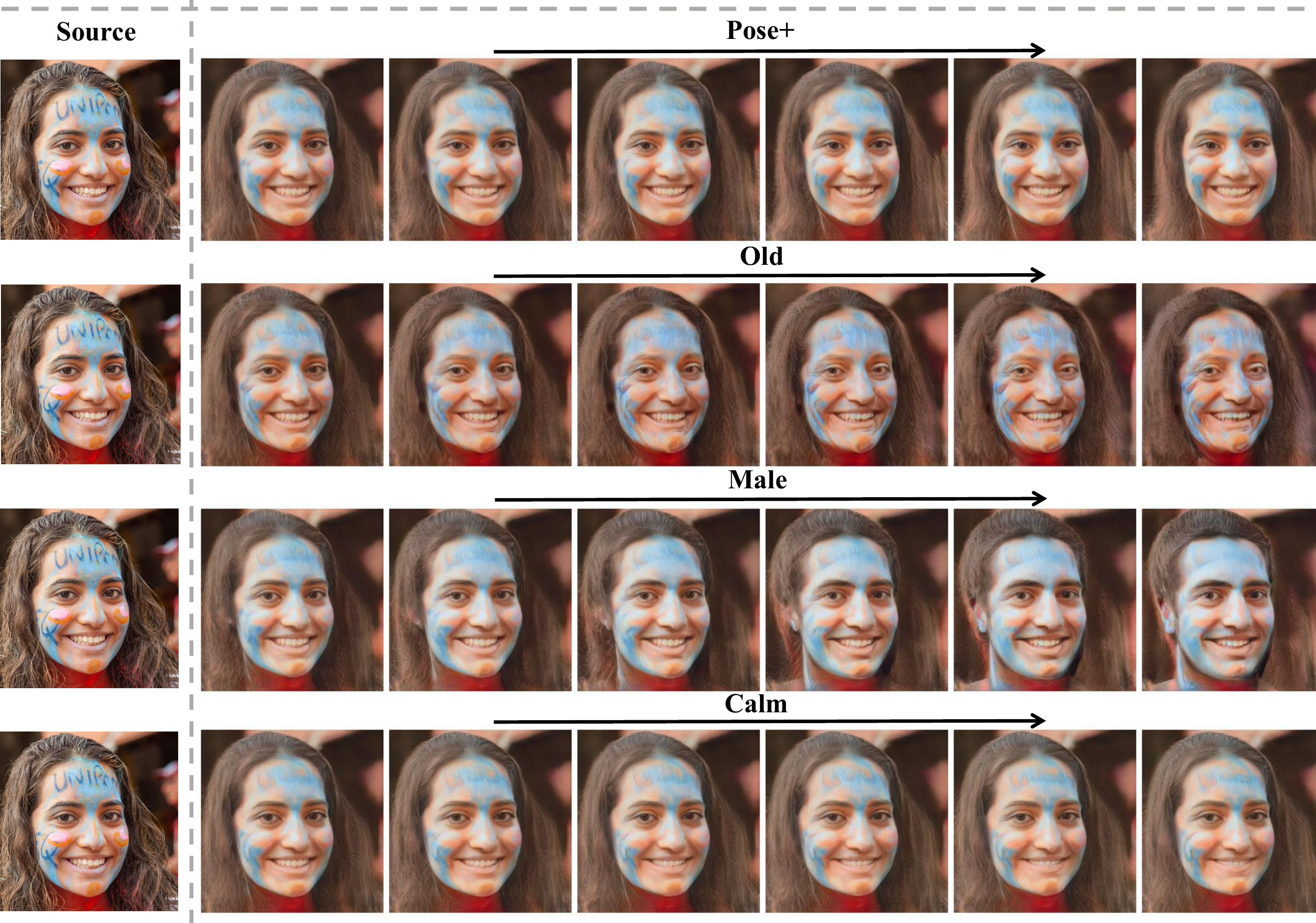}
    \caption{Qualitative results of our method on real image editing. The images are taken from FFHQ.}
    \label{real_3}
\end{figure*}

We show more real-image editing results of our method on different attributes (including pose) in Figs.~\ref{real_1} and~\ref{real_3}. We can see that our method can obtain high-quality disentangled results when editing various face attributes. It implies the effectiveness of our method.

\subsection{More Results on Multiple Condition Attributes}
\begin{figure*}[t]
    \centering
    \includegraphics[width=1\textwidth]{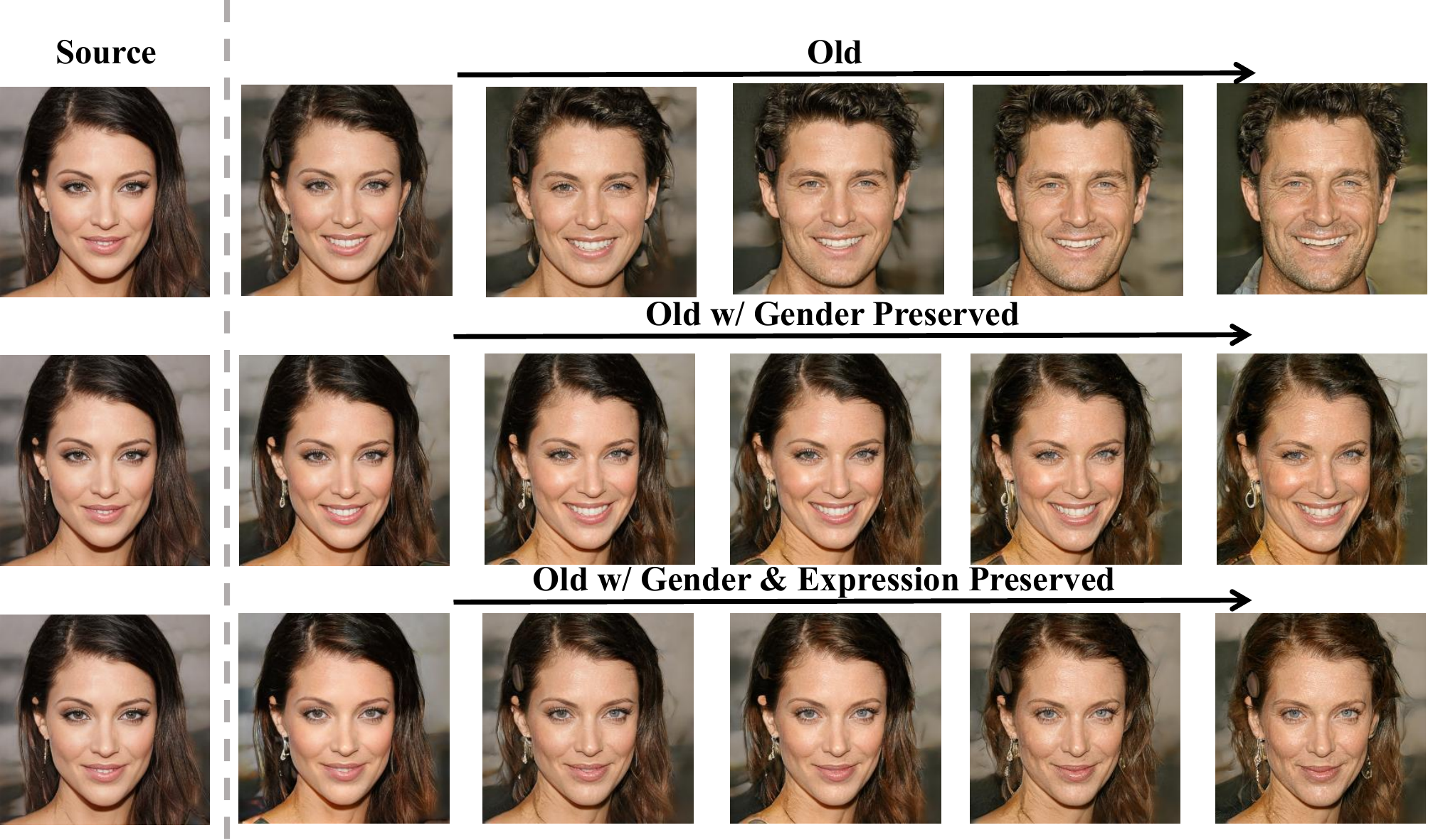}
    \includegraphics[width=1\textwidth]{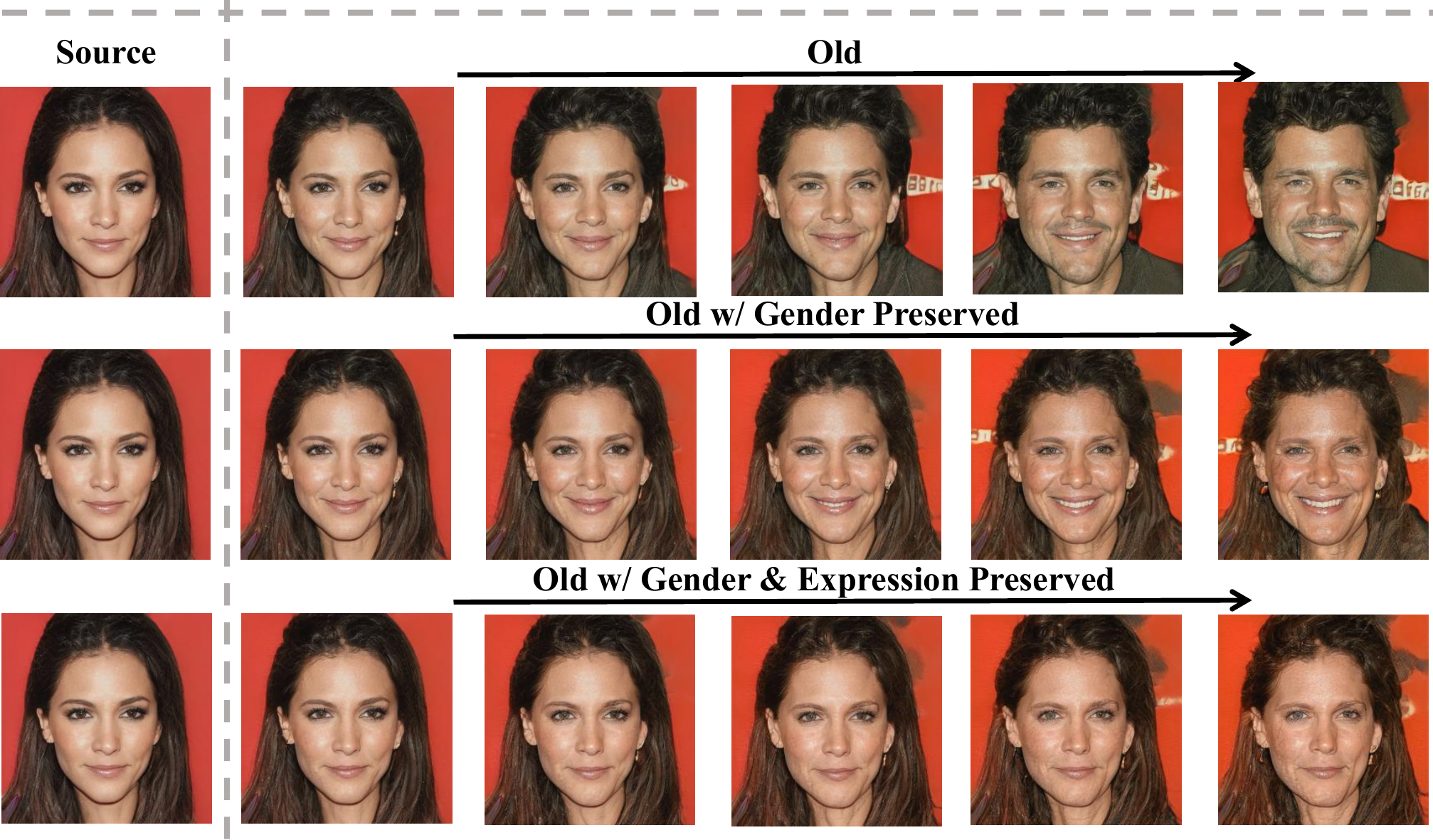}
    \caption{Qualitative results of our method on multiple condition attributes. }
    \label{multi_cond}
\end{figure*}

In Fig~\ref{multi_cond}, we show the facial editing results of our method with a different number of condition attributes. 
%Here, the condition attributes are manually specified according to the strategy we have discussed in Sec.~\ref{sec}.
It can be observed that our method is able to handle complex disentanglement between face attributes by adding more condition attributes into the conditional manipulation operation using the strategy presented in \cite{shen2020interpreting}.

\subsection{More Ablation Study on Control Factors.}
\begin{figure*}[t]
    \centering
    \includegraphics[width=1\textwidth]{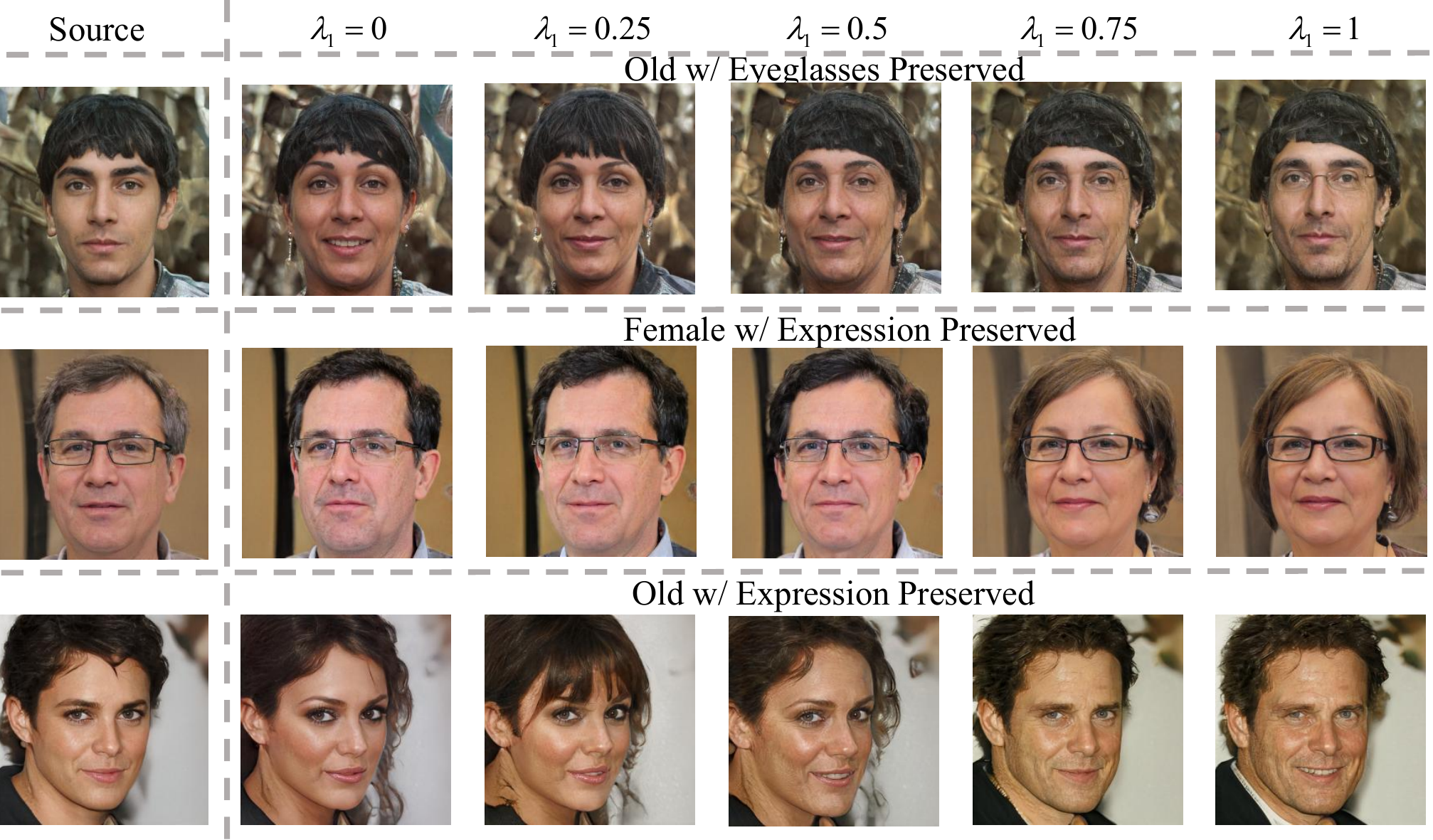}
    \caption{Qualitative ablation study on $\lambda_1$ (we choose $\lambda_1=0.75$ in this paper based on the $DT$ metric). }
    \label{control_factor_1}
\end{figure*}

\begin{figure*}[t]
    \centering
    \includegraphics[width=1\textwidth]{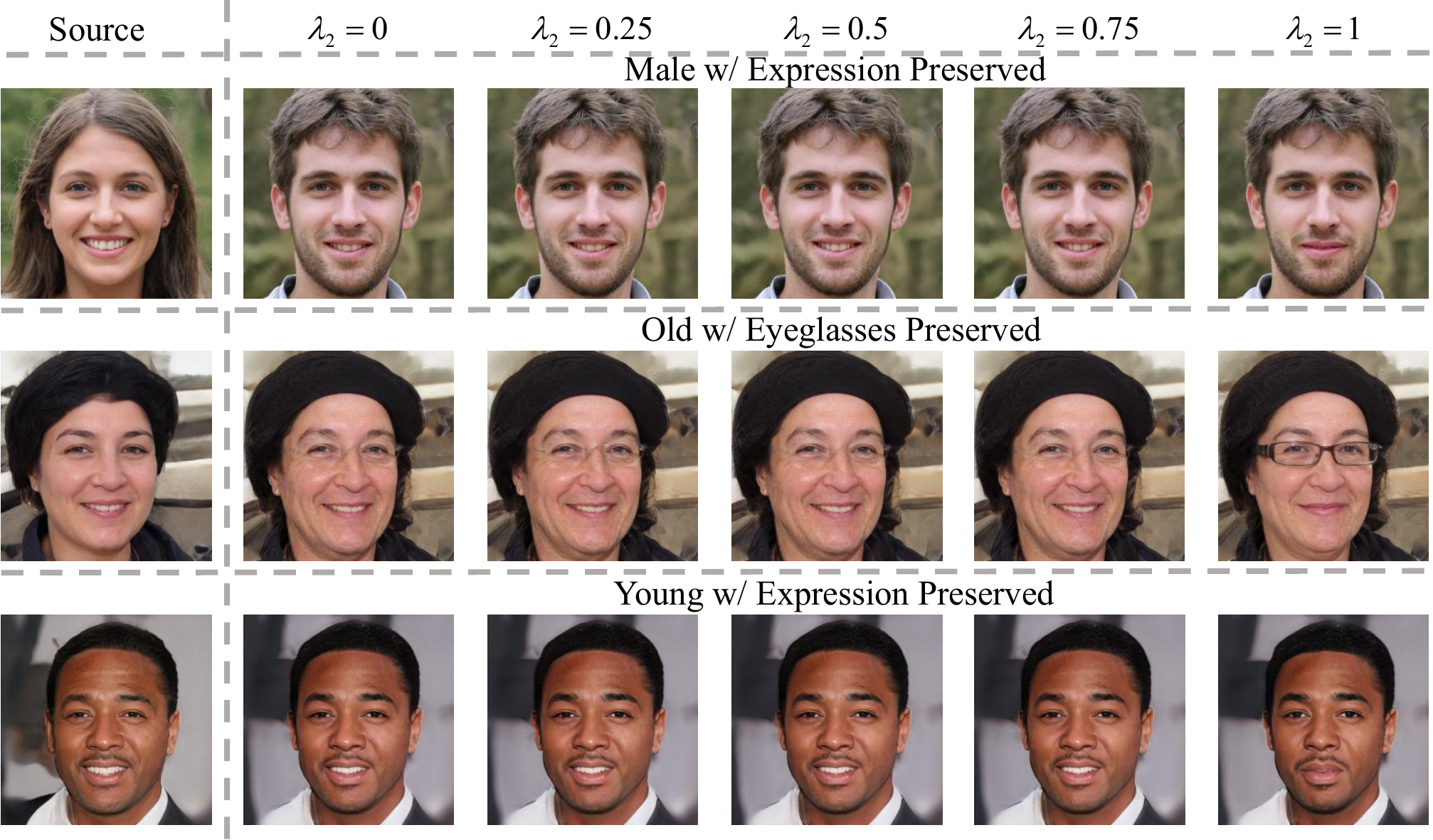}
    \caption{Qualitative ablation study on $\lambda_2$ (we choose $\lambda_2=0$ in this paper based on the $DT$ metric).}
    \label{control_factor_2}
\end{figure*}

In Figs.~\ref{control_factor_1} and~\ref{control_factor_2}, we show the qualitative ablation results of varying $\lambda$ in the primal and condition directions, respectively. 
We can see that \emph{i}) our method would fail to change the desired primal attribute if we abandon the attribute-level information, and \emph{ii}) our method can effectively obtain disentangled results when adding instance information, which is consistent with the conclusion in the main paper.

\subsection{More Comparison with the State-of-the-Arts}
\begin{figure*}[t]
    \centering
    \includegraphics[width=0.9\textwidth]{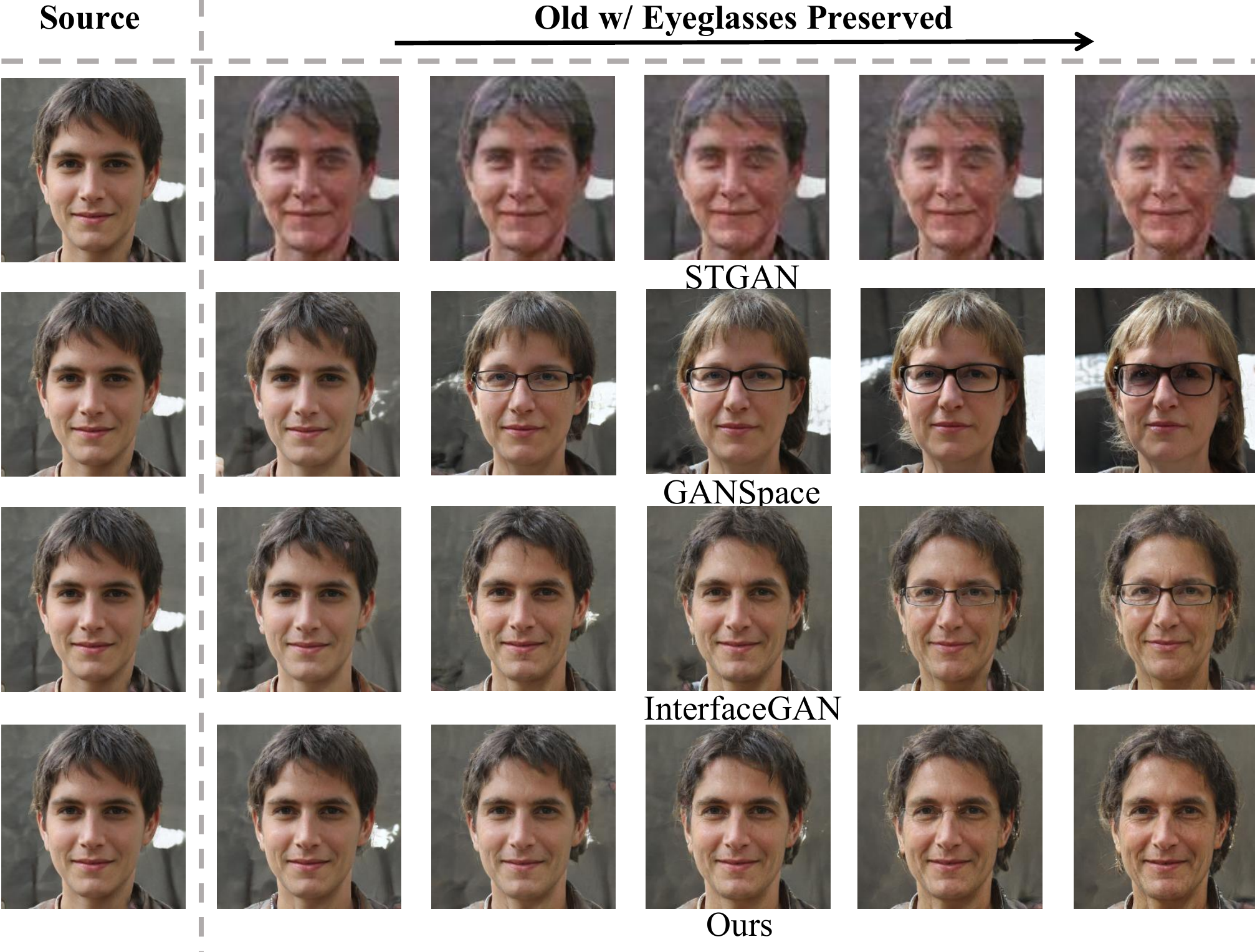}
    \includegraphics[width=0.9\textwidth]{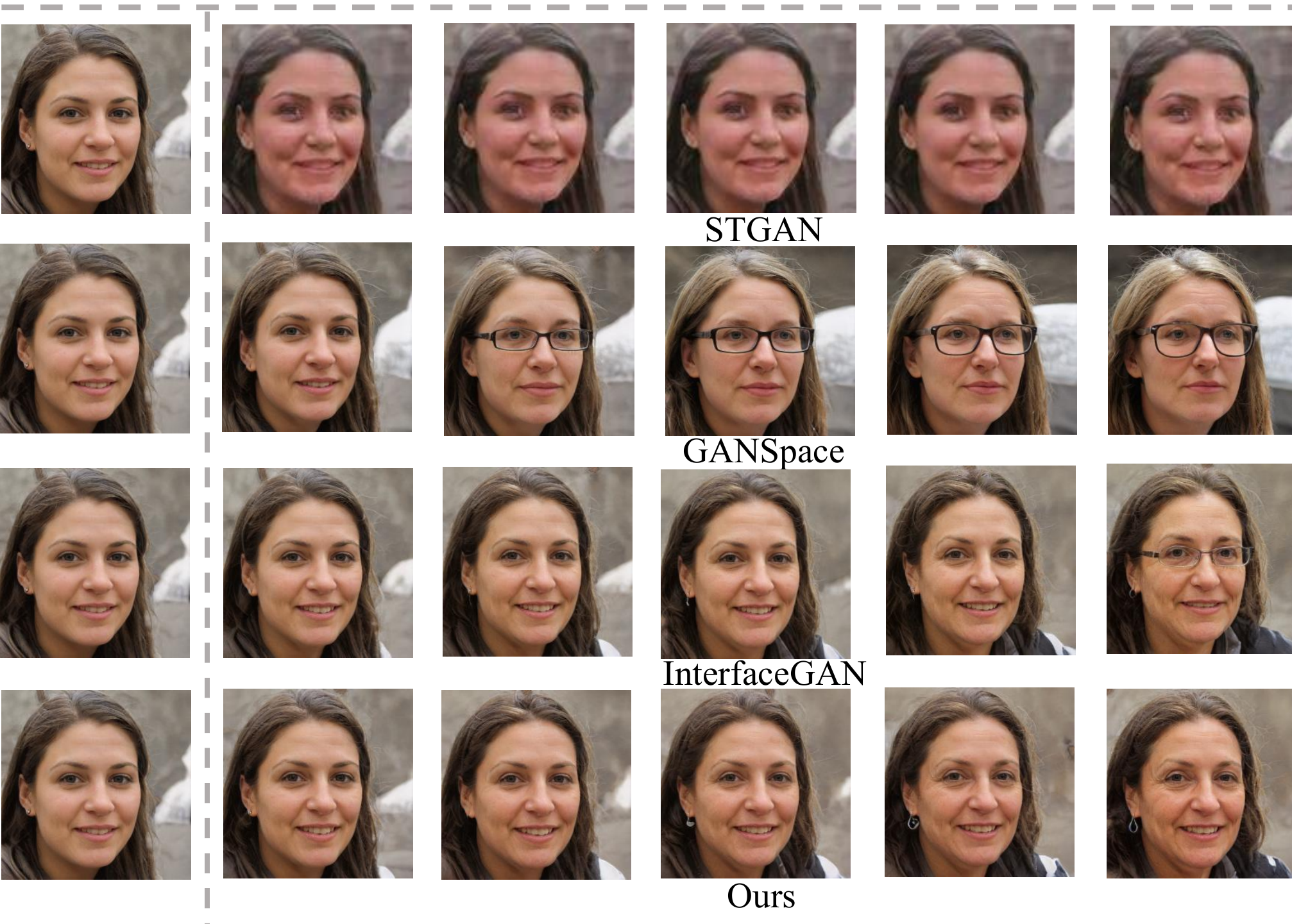}
    \caption{More qualitative comparison with the state-of-the-art %of disentangled facial editing between our method and the competitors 
    on GAN-generated images.}
    \label{compare_1}
\end{figure*}

\begin{figure*}[t]
    \centering
    \includegraphics[width=0.9\textwidth]{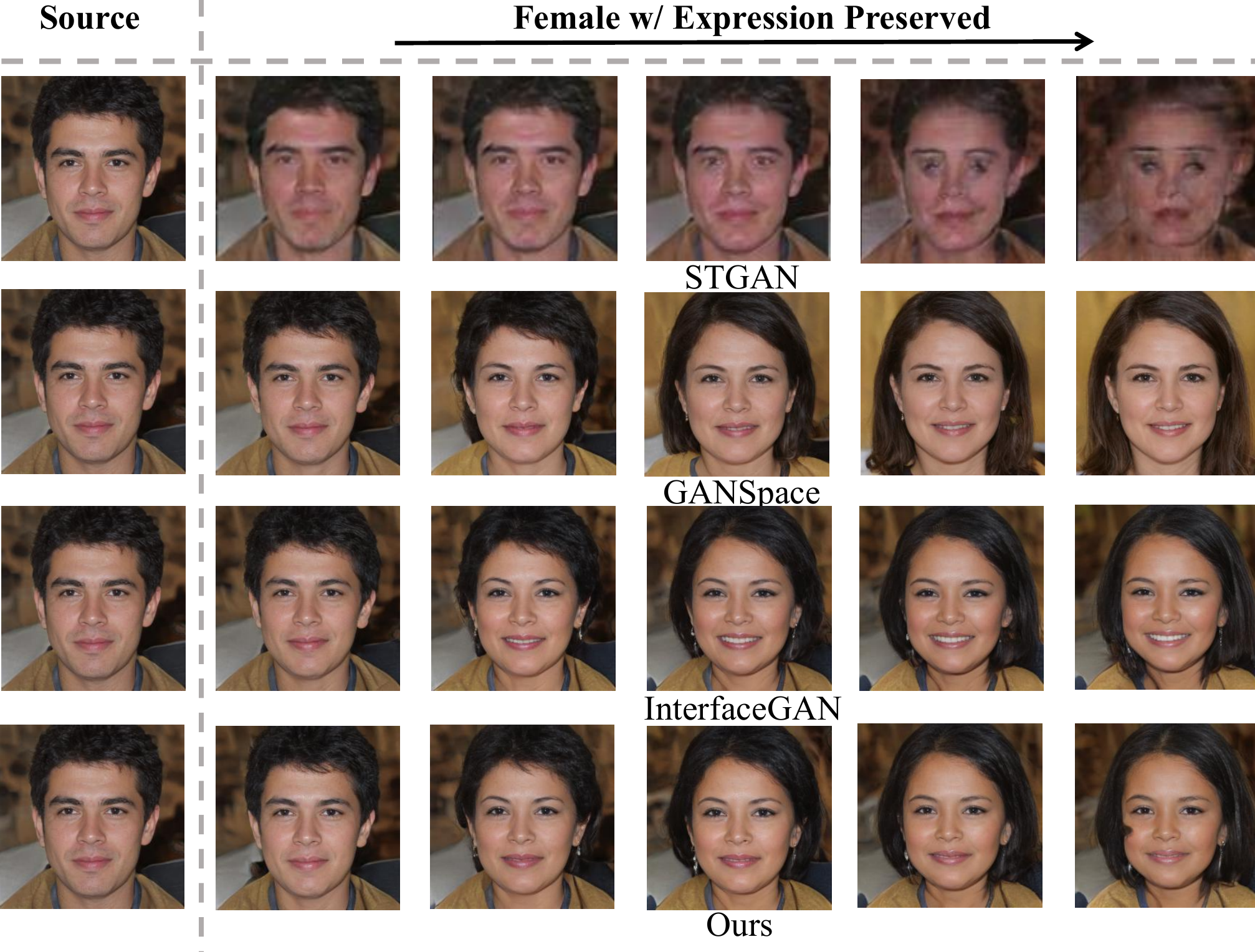}
    \includegraphics[width=0.9\textwidth]{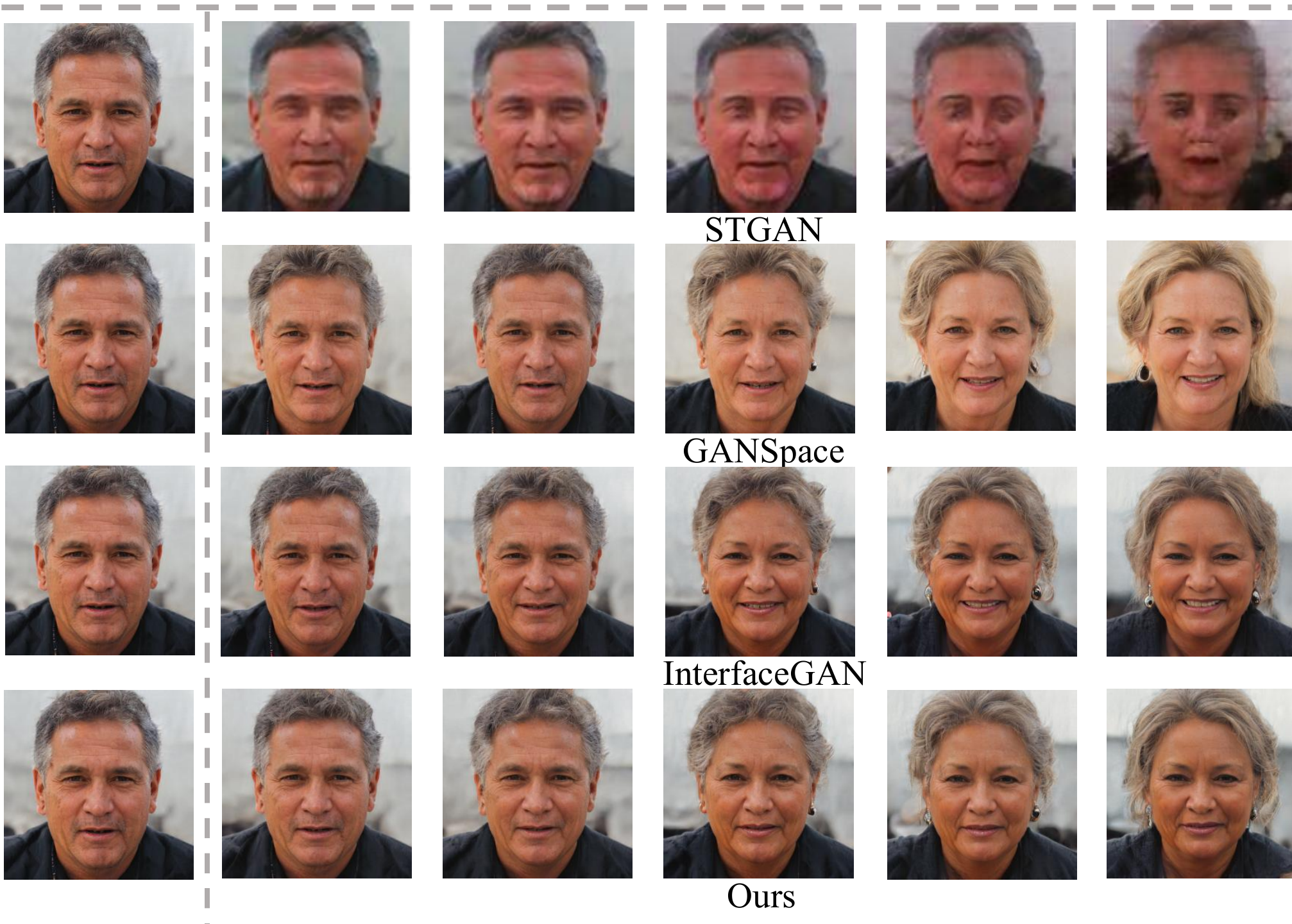}
    \caption{More qualitative comparison with the state-of-the-art %of disentangled facial editing between our method and the competitors 
    on GAN-generated images.}
    \label{compare_2} 
\end{figure*}

\begin{figure*}[t]
    \centering
    \includegraphics[width=0.9\textwidth]{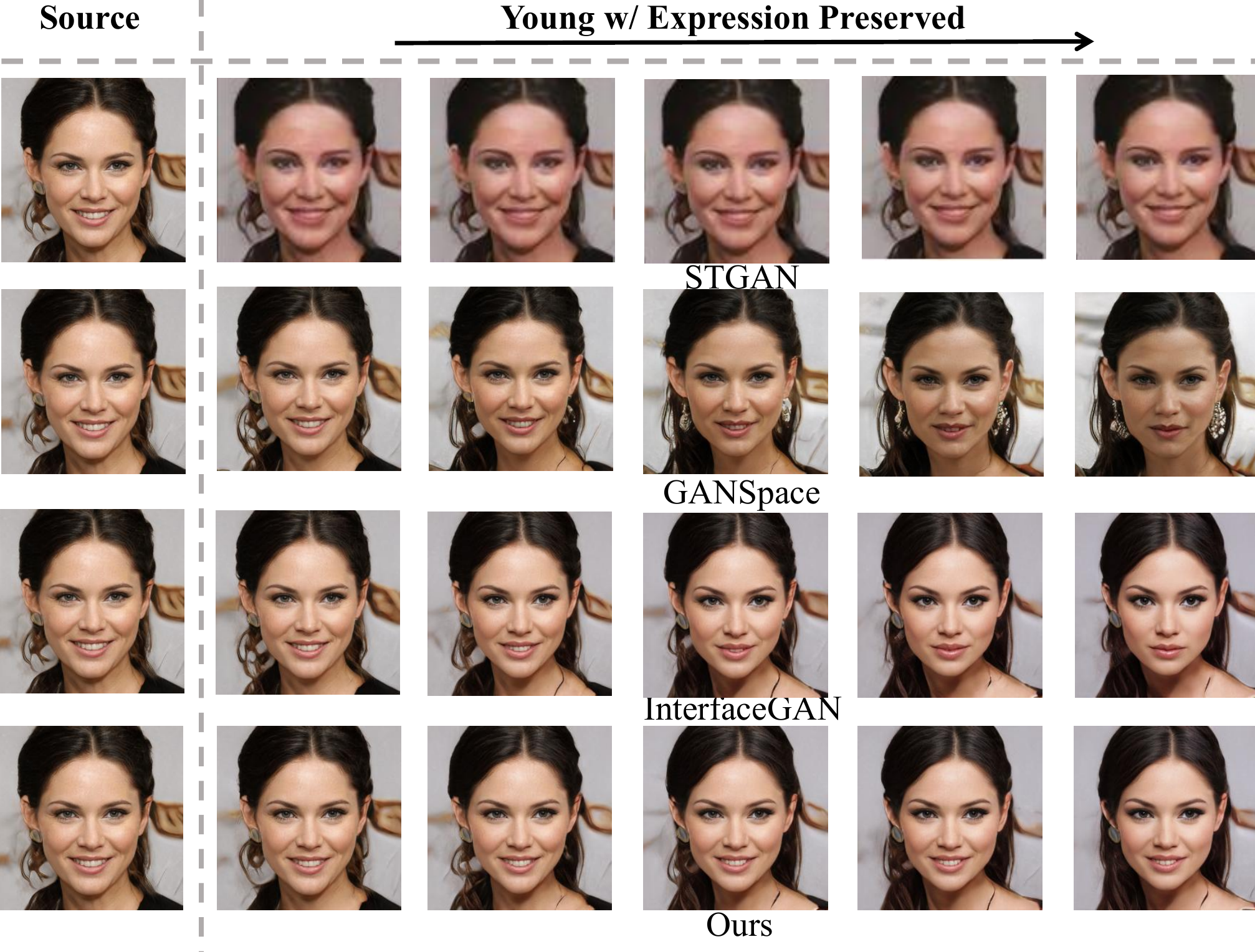}
    \includegraphics[width=0.9\textwidth]{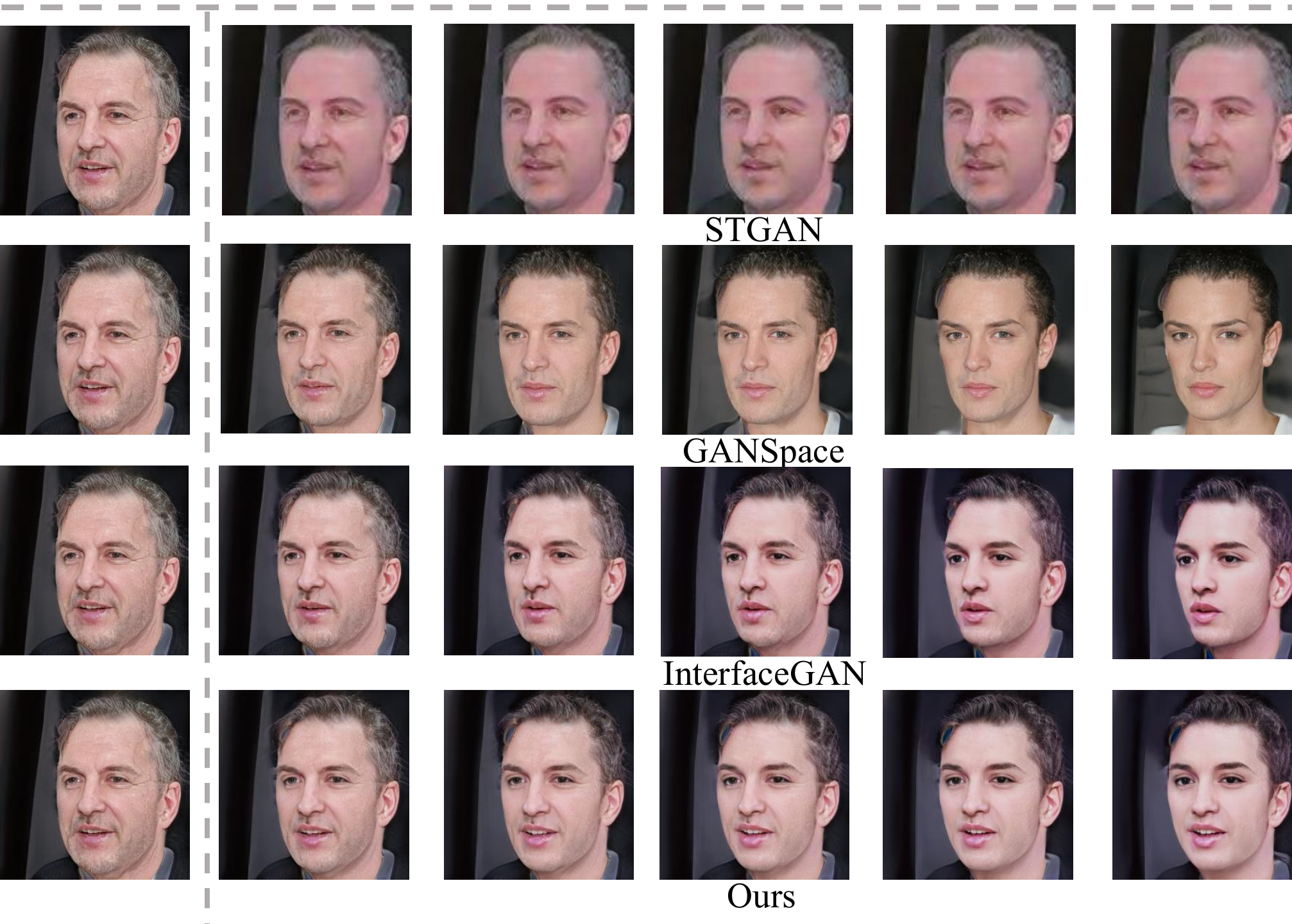}
    \caption{More qualitative comparison with the  state-of-the-art %of disentangled facial editing between our method and the competitors 
    on GAN-generated images.}
    \label{compare_3}
\end{figure*}

Here, we provide more qualitative results compared to the state-of-the-art methods, including the semantic direction based methods (InterfaceGAN~\cite{shen2020interpreting} and GANSpace~\cite{h2020ganspace}), and the conditional-GAN based method~\cite{liu2019stgan} on GAN-generated images.
%They utilize the same settings mentioned in the main paper, \emph{i.e.}
We adopt the conditional manipulation setup~\cite{shen2020interpreting} for the semantic direction based methods, and directly send the source image to the conditional-GAN based method, as in the main paper. 
Besides, the results in Figs.~\ref{compare_1},~\ref{compare_2}, and~\ref{compare_3} show that our method performs better than the other competitors.

\subsection{More Comparison with InterfaceGAN on attribute-level direction}
\begin{figure*}[t]
    \centering
    \includegraphics[width=1\textwidth]{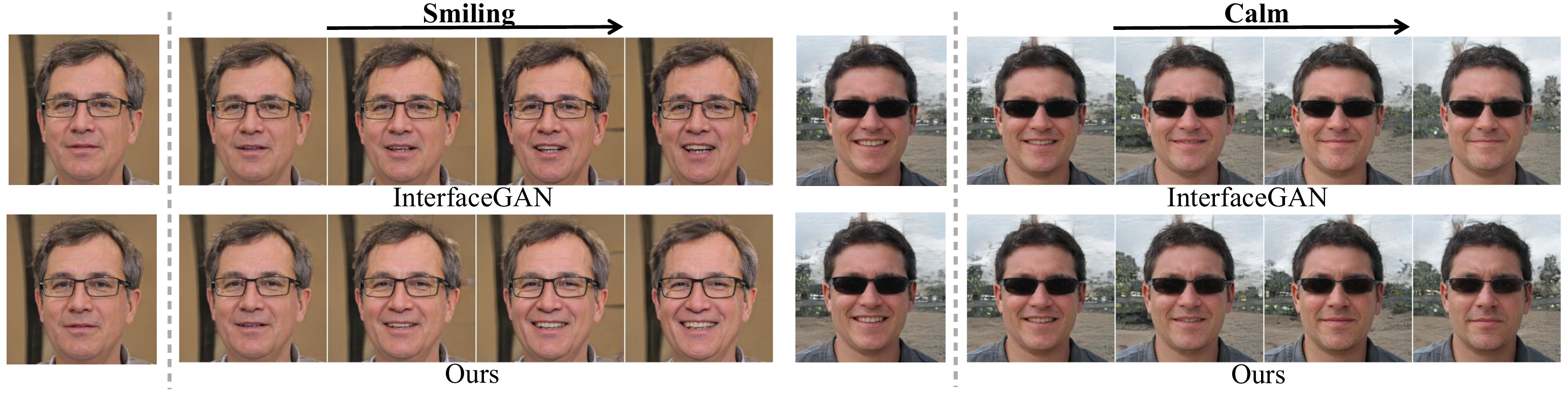}
    \includegraphics[width=1\textwidth]{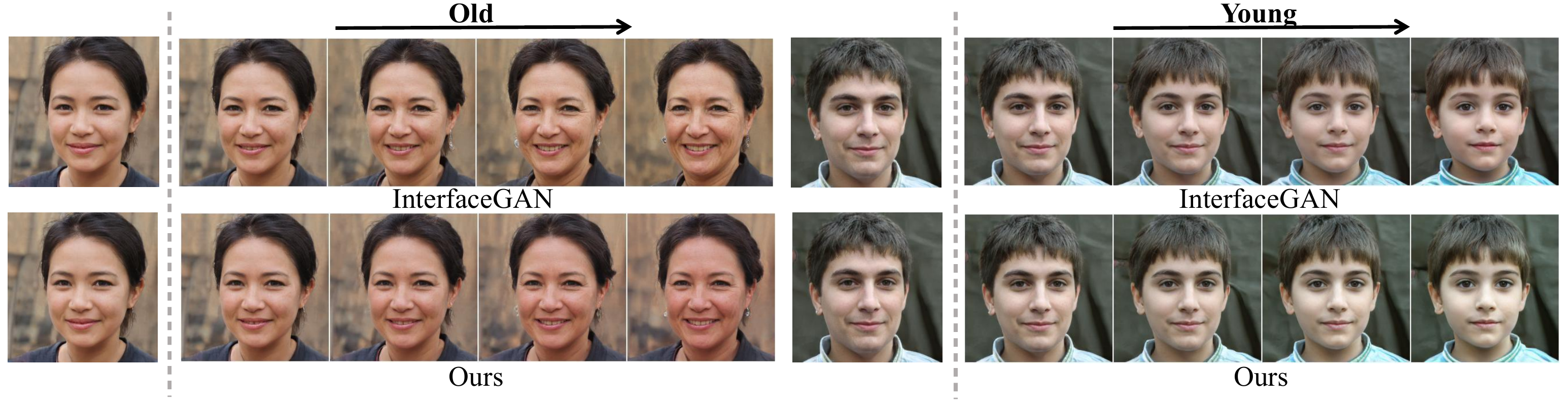}
    \includegraphics[width=1\textwidth]{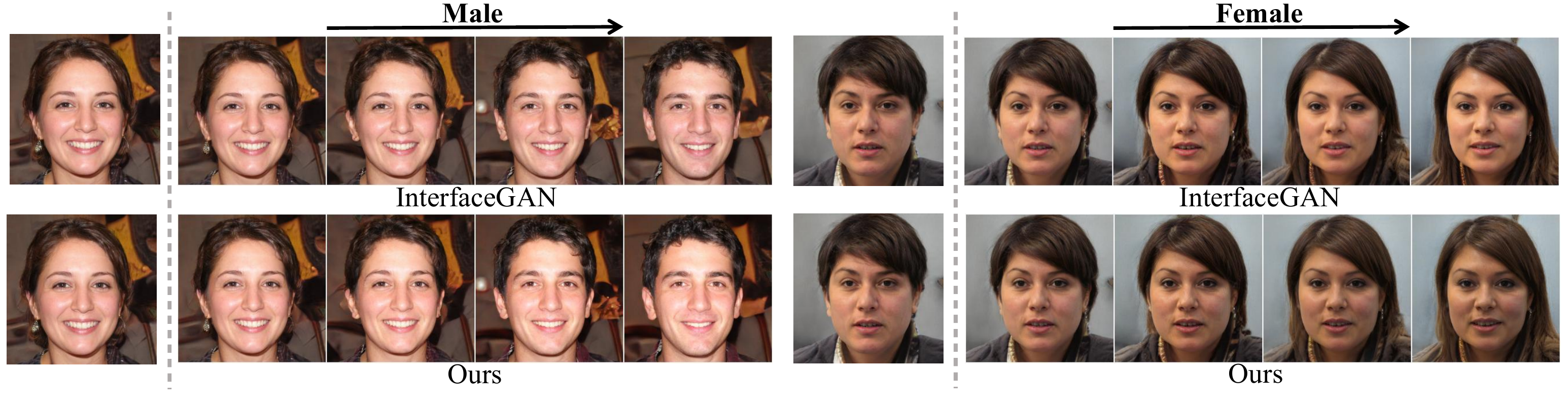}
    \includegraphics[width=1\textwidth]{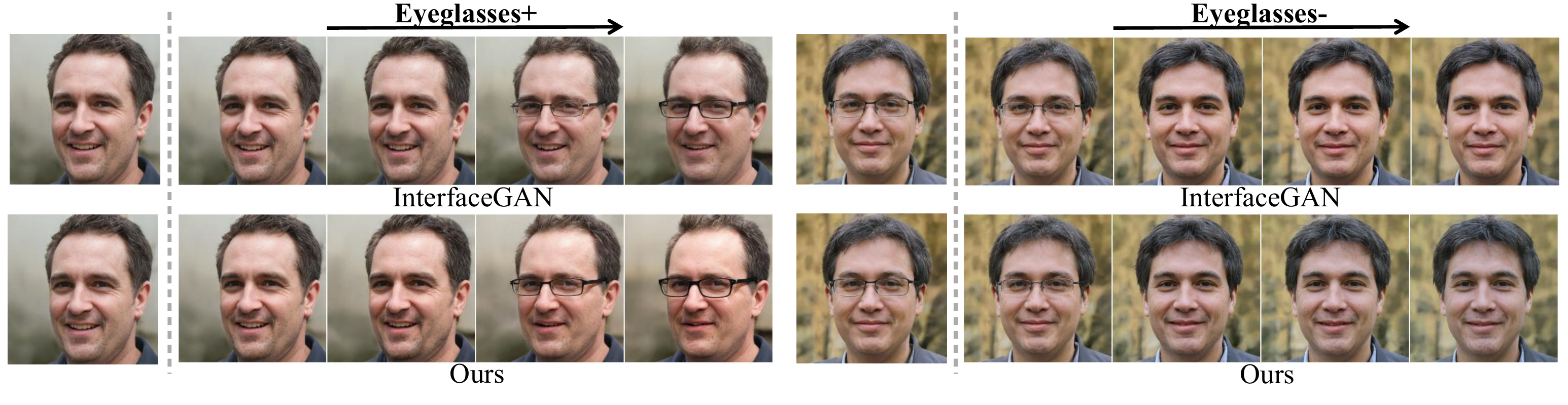}
    \includegraphics[width=1\textwidth]{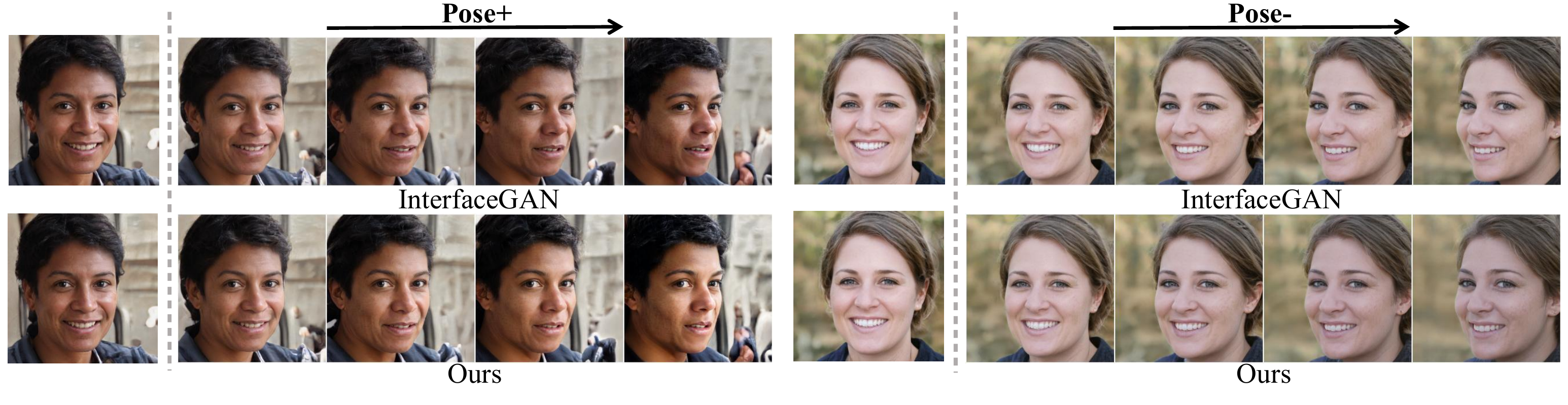}
    \caption{Qualitative comparison of face image editing using the attribute-level directions computed by our method and %the counterpart computed by
    InterfaceGAN.}
    \label{attr_level_compare_to_interfacegan}
\end{figure*}

In the main paper, we point out that the attribute-level direction computed by our method leads to similar editing quality with the counterpart computed by InterfaceGAN~\cite{shen2020interpreting}. 
Here, we show qualitative comparison in  Fig.~\ref{attr_level_compare_to_interfacegan} and demonstrate their similar efficacy.
Note that our process is easier to implement, as we eliminate the need for training SVMs.
%Note that the advantage of implementation efficiency is just a by-product of our method. 
Our main contribution is to use the instance-aware semantic direction for face attribute editing.   

\section{More Details of the Incremental Updating Scheme}
In the main paper, we propose an incremental instance-aware direction
search scheme, instead of keep using the instance aware direction $\hat{d}_X(z)$ of the original latent code $z$ when editing.  
%The intuition behind is straightforward. 
We treat the latent codes formed in the editing process (\emph{e.g.} $z+k\hat{d}_X(z)$, where $k$ is the step size) as different instances and use their instance-aware direction to edit them.
For example, we denote $z'=z+k\hat{d}_X(z)$ and adopt $\hat{d}_X(z')$ to edit $z'$ instead of directly using $\hat{d}_X(z)$.
Note that the incremental updating scheme makes no difference if we use the attribute-level direction alone in the editing process.

\section{Discussion about Condition Attribute Selection}\label{sec}
In practice, one can freely select one or multiple condition attributes for primal attribute editing, depending on the targeted editing goals. In general, to achieve accurate and disentangled editing for a primal attribute, selection of condition attribute is closely related to the data distribution of GAN's training set. For example, elder people tend to wear eyeglasses in FFHQ~\cite{karras2020a}, so age and eyeglass attributes are likely to be entangled with each other in the latent space. One can also follow this strategy for condition attribute selection.

\section{Limitations}
We have demonstrated that our method is effective for disentangled face attribute editing. It still has some limitations of the semantic direction based facial editing algorithm.
For example, our method cannot model non-binary attributes (\emph{e.g.} hair color and hairstyle), as it needs a binary classification boundary for facial editing.
%On the other hand, our method needs to manually select the condition attribute in the conditional manipulation~\cite{shen2020interpreting} setup.
In addition, we observed that the image editing quality on embedded latent code obtained by GAN inversion approaches~\cite{abdal2019image2stylegan,abdal2020image2stylegan} is slightly lower than random latent codes from the GAN generator. A recent work from \cite{zhu2020in} proposed to eliminate the above performance gap by adding a regularization term to encourage the embedded latent code to be semantically meaningful, which we will explore in our future work.

% \begin{figure*}[t]
%     \centering
%     \includegraphics[width=0.95\textwidth, trim=0 0 190 10, clip]{supp_img/real_3.pdf}
%     \includegraphics[width=0.95\textwidth, trim=0 0 190 0, clip]{supp_img/real_4.pdf}
%     \caption{More qualitative results of our method on real image editing.}
%     \label{real_2}
% \end{figure*}

\bibliographystyle{named}
\bibliography{ijcai21}

\end{document}